\documentclass[runningheads]{llncs}
\usepackage{amsmath,amssymb} 
\usepackage{algorithm}
\usepackage[noend]{algpseudocode}
\usepackage{multirow}
\usepackage{color}
\usepackage{xspace}
\usepackage{bm}
\usepackage{graphicx}
\usepackage{subcaption}
\usepackage{afterpage}
\usepackage{booktabs,tabularx}
\newcommand{\vect}[1]{\boldsymbol{#1}}
\newcommand{\bea}{\begin{eqnarray}} 
\newcommand{\eea}{\end{eqnarray}}
\newcommand{\be}{\begin{equation}} 
\newcommand{\ee}{\end{equation}}

\begin{document}
\pagestyle{headings}
\mainmatter
\def\ECCV18SubNumber{***}  

\title{Structural Consistency and Controllability for Diverse Colorization} 

\titlerunning{Structural Consistency and Controllability for Diverse Colorization}

%
\titlerunning{Structural Consistency and Controllability for Diverse Colorization}
%


\newcommand*{\ie}{i.e.\@\xspace}
\newcommand*{\eg}{e.g.\@\xspace}
\newcommand{\etal}{\textit{et al}.}
\newcommand{\cD}{\mathcal{D}}

\author{Safa Messaoud \and
David Forsyth\and
Alexander G. Schwing}

\institute{University of Illinois at Urbana-Champaign, USA}

\maketitle

\begin{abstract} Colorizing a given gray-level image is an important task in the media and advertising industry. Due to the  ambiguity inherent to colorization (many shades are often plausible), 
recent approaches started to explicitly model diversity. 
 However, one of the most obvious artifacts, structural inconsistency, is rarely considered by existing methods which  predict chrominance independently for every pixel. To address this issue, we develop a conditional random field based variational auto-encoder formulation which is able to achieve diversity while taking into account structural consistency. Moreover, we introduce a controllability mechanism that can incorporate external constraints from diverse sources including a user interface. Compared to existing baselines, we demonstrate that our method obtains more diverse and globally consistent colorizations on the LFW, LSUN-Church and ILSVRC-2015 datasets.

\end{abstract}

\section{Introduction}

Colorization of images requires to predict the two missing channels of a provided gray-level input. Similar to other computer vision tasks like monocular depth-prediction or semantic segmentation, colorization is ill-posed. However, unlike the aforementioned tasks, colorization is also ambiguous, \ie, many different colorizations are perfectly plausible. For instance, differently colored shirts or cars are very reasonable, while there is certainly less diversity in shades of  fa\c{c}ades. 
Capturing these subtleties is a non-trivial problem.

Early work on colorization was therefore interactive, requiring some reference color image or scribbles~\cite{welsh2002,levin2004,chia2011,gupta2012,ironi2005,morimoto2009}. 
To automate the process, classical methods formulated the task as a prediction problem~\cite{charpiat2008,deshpande2015}, using  datasets of limited sizes. More recent deep learning methods were shown to capture more intricate color properties on larger datasets~\cite{cheng2015,iizuka2016,larsson2016,zhang2016,varga2017,zhang2017}. However, all those methods have in common that they only produce a single colorization for a given gray-level image. 
Hence, the ambiguity and multi-modality are often not modeled adequately. To this end, even more recently, diverse output space distributions for colorization were described using generative modeling techniques such as variational auto-encoders~\cite{deshpande2016}, generative adversarial nets~\cite{isola2017}, or auto-regressive models~\cite{guadarrama2017,royer2017}.

\begin{figure}[t!]
\centering
\includegraphics[width=1.\linewidth]{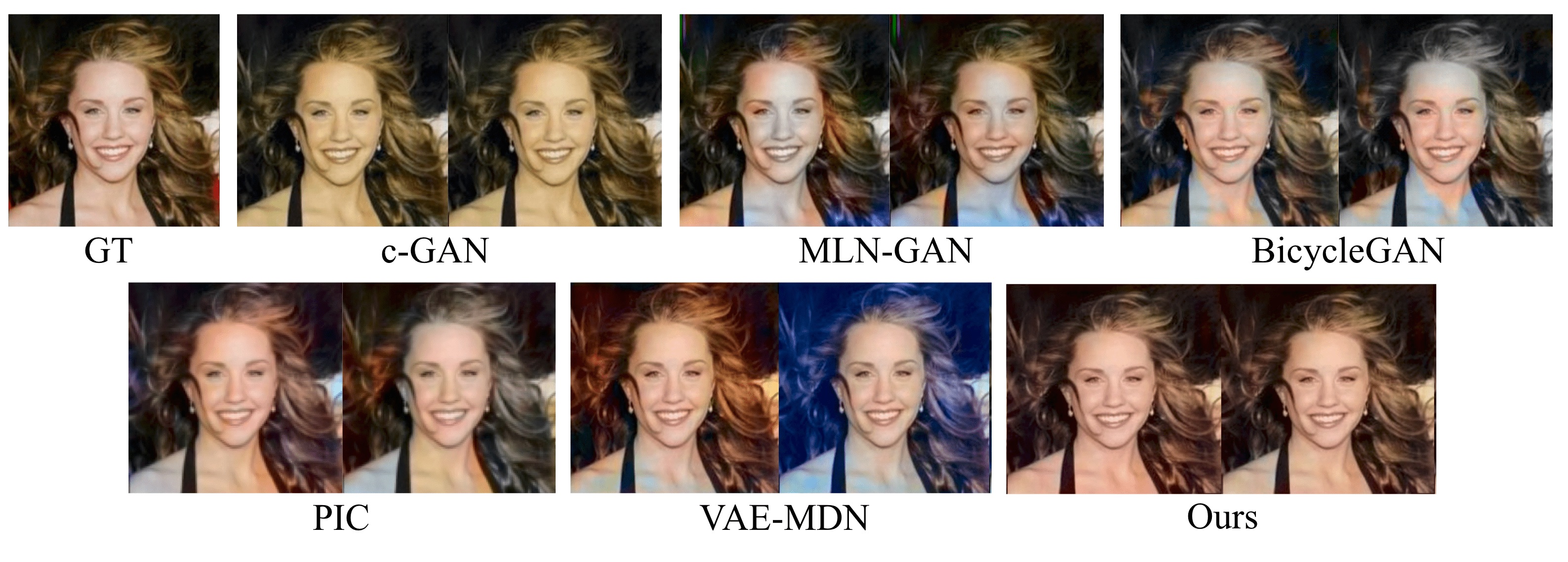}

\caption{{\small Diverse colorizations of the ground truth (GT) generated  by c-GAN~\cite{isola2017}, MLN-GAN~\cite{Cao2017}, BicycleGAN~\cite{zhu2017toward}, PIC~\cite{royer2017}, VAE-MDN~\cite{deshpande2016} and our approach. }}
\label{fig:first_fig}

\end{figure}

While approaches based on generative techniques can produce diverse colorizations 
by capturing a dataset distribution, they often  lack structural consistency, \eg, parts of a shirt differ in  color or the car is speckled. Inconsistencies are due to the fact that structural coherence is only encouraged implicitly when using deep net based generative methods. For example, in results obtained from~\cite{deshpande2016,isola2017,royer2017,Cao2017,zhu2017toward} illustrated in Fig.~\ref{fig:first_fig}, the color of the shoulder and  neck differ as these models are sensitive to occlusion. 
In addition, existing diverse colorization techniques also often lack a form of controllability permitting to interfere while maintaining structural consistency. 

To address both consistency and controllability, 
our developed method enhances the output space of variational auto-encoders~\cite{kingma2013} with a Gaussian Markov random field formulation. Our developed approach, which we train in an end-to-end manner, enables explicit modeling of the structural relationship between multiple pixels in an image. Beyond learning the structural consistency between pixels, we also develop a control mechanism which  incorporates external constraints. This enables a user to interact with the generative process using color stokes. 
We illustrate visually appealing results on the Labelled Faces in the Wild (LFW)~\cite{learned2016labeled}, LSUN-Church~\cite{yu2015lsun} and ILSVRC-2015~\cite{russakovsky2015imagenet} datasets and assess the photo-realism aspect with a user study.

\section{Related Work}

As mentioned before, we develop a colorization technique which enhances variational auto-encoders with Gaussian Markov random fields. Before discussing the details, we review the three areas of colorization, Gaussian Markov random fields and variational auto-encoders subsequently.

\noindent{\bf Colorization:}  
Early colorization methods rely on user-interaction in the form of a reference image or scribbles~\cite{welsh2002,levin2004,chia2011,gupta2012,ironi2005,morimoto2009}. First attempts to automate the colorization process~\cite{charpiat2008} rely on classifiers trained on datasets containing a few tens to a few thousands of images. Naturally, recent deep net based methods scaled to much larger datasets containing millions of images~\cite{cheng2015,iizuka2016,larsson2016,zhang2016,varga2017,zhang2017,varga2017twin}.
All these methods operate on a provided intensity field and produce a single color image which doesn't embrace the ambiguity of the task. 

To address ambiguity, Royer \etal~\cite{royer2017} use a PixelCNN~\cite{van2016conditional} to learn a conditional  model $p(\vect x|\vect g)$ of the color field $\vect x$ given the gray-level image $\vect g$, and draw multiple samples from this distribution to obtain different colorizations. 
In addition to compelling results, failure modes are reported due to ignored complex long-range pixel interactions, \eg, if an object is split due to occlusion. Similarly,~\cite{guadarrama2017} uses PixelCNNs to learn multiple embeddings $\vect{z}$ of the gray-level image, 
before a convolutional refinement network is trained to obtain the final image. Note that in this case, instead of learning $p(\vect{x}|\vect{g})$ directly, the color field $\vect{x}$ is represented by a low dimensional embedding $\vect z$. Although, the aforementioned PixelCNN based approaches yield diverse colorization, they lack large scale spatial coherence and are prohibitively slow due to the auto-regressive, \ie, sequential, nature of the model.

Another conditional latent variable approach for diverse colorization was proposed by Deshpande \etal~\cite{deshpande2016}. The authors train a variational auto-encoder to produce a low dimensional embedding of the color field. Then, a Mixture Density Network (MDN)~\cite{bishop1994mixture} is used to learn a multi-modal distribution $p(\vect{z}|\vect{g})$ over the latent codes. Latent samples are afterwards converted to multiple color fields using a decoder. This approach offers an efficient sampling mechanism. However, the output is often speckled because colors are sampled independently for each pixel. 

Beyond the aforementioned probabilistic formulations, conditional generative adversarial networks~\cite{isola2017} have been used to produce diverse colorizations. However, mode collapse, which results in the model producing one color version of the gray-level image, is a frequent concern in addition to consistency. This is mainly due to the generator learning to largely ignore the random noise vector when conditioned on a relevant context.~\cite{Cao2017} addresses the former issue by concatenating the input noise channel with several convolutional layers of the generator. A second solution is proposed by~\cite{zhu2017toward}, where the connection between the output and latent code is encouraged to be invertible to avoid many to one mappings. These models show compelling results when tested on datasets with strong alignment  between the samples, \eg, the LSUN bedroom dataset~\cite{yu2015lsun} in~\cite{Cao2017} and image-to-image translation datasets~\cite{Laffont14,cordts2016cityscapes,isola2017image,yu2014fine,zhu2016generative} in~\cite{zhu2017toward}. We will demonstrate in Sec.~\ref{sec:exp} that they lack global consistency on more complex datasets.

In contrast to the aforementioned formulations, we address both diversity and global structural consistency requirements while ensuring computational efficiency. To this end we formulate the colorization task by augmenting variational auto-encoder models with Gaussian Conditional Random Fields (G-CRFs). Using this approach, beyond modeling a structured output space distribution, controllability of the colorization process is natural.


\noindent{\bf Gaussian Conditional Markov Random Field:} Markov random fields~\cite{kindermann1980markov} and their conditional counter-part are a compelling tool to model correlations between variables. 
Theoretically, they are hence a good match for colorization tasks where we are interested in reasoning about color dependencies between different pixels. However, inference of the most likely configuration in classical Markov random fields defined over large output spaces is computationally demanding~\cite{SchwingCVPR2011a,SchwingNIPS2012,SchwingICML2014,MeshiNIPS2017} and only tractable in a few special cases.

Gaussian Markov random fields~\cite{rue2008gaussian} represent one of those cases which permit efficient and exact inference. They model the joint distribution of the data, \eg, the pixel values of the two color channels of an image as a multi-variate Gaussian density. Gaussian Markov random fields have been used in the past for different computer vision applications including semantic segmentation~\cite{vemulapalli2016gaussian,chandra2016,chandra2016fast}, human part segmentation and saliency estimation~\cite{chandra2016,chandra2016fast}, image labeling~\cite{jancsary2012regression} and image denoising~\cite{tappen2007learning,vemulapalli2016deep}. A sparse Gaussian conditional random field trained with a LEARCH framework has been proposed for colorization in~\cite{deshpande2015}. Different from this approach, we use a fully connected Gaussian conditional random field and learn its parameters end-to-end with a deep net. 
Beyond structural consistency, our goal is to jointly model the ambiguity which is an inherent part of the colorization task. To this end we make use of variational auto-encoders.

\noindent{\bf Variational Auto-Encoders:} Variational auto-encoders (VAEs)~\cite{kingma2013} and  conditional variants~\cite{sohn2015learning}, \ie, conditional VAEs (CVAEs), have been used to model ambiguity in a variety of tasks~\cite{WangNIPS2017,JainZhangCVPR2017}. They are based on the manifold assumption stating that a high-dimensional data point  $\vect x$, such as a color image, can be modeled based on a low-dimensional embedding $\vect z$ and some auxiliary data $\vect g$, such as a gray-level image. Formally, existence of a low-dimensional embedding space and a transformation via the conditional $p_\theta(\vect x|\vect z, \vect g)$ is assumed. Given a dataset $\cD$ containing pairs of conditioning information $\vect g$ and desired output $\vect x$, \ie, given $\cD = \{(\vect g, \vect x)\}$, CVAEs formulate maximization of the conditional log-likelihood $\ln p_\theta(\vect x|\vect g)$, parameterized by $\theta$, by considering the following identity: 
\bea
&&\hspace{-0.7cm}\ln p_\theta(\vect x|\vect g) - D_{\operatorname{KL}}(q_\phi(\vect z|\vect x, \vect g), p_\theta(\vect z|\vect x, \vect g)) =\label{eq:CVAE}\\
&&\hspace{-0.5cm} -D_{\operatorname{KL}}(q_\phi(\vect z|\vect x, \vect g), p(\vect z|\vect g)) + \mathbb{E}_{q_\phi(\vect z|\vect x, \vect g)}[\ln p_\theta(\vect x|\vect g, \vect z)].\nonumber
\eea
Hereby, $D_{\operatorname{KL}}(\cdot, \cdot)$ denotes the Kullback-Leibler (KL) divergence between two distributions, and $q_\phi(\vect z|\vect x, \vect g)$ is used to approximate the intractable posterior $p_\theta(\vect z|\vect x, \vect g)$ of a deep net which models the conditional $p_\theta(\vect x|\vect g, \vect z)$. The approximation of the posterior, \ie, $q_\phi(\vect z|\vect x, \vect g)$, is referred to as the encoder, while the deep net used for reconstruction, \ie, for modeling the conditional $p_\theta(\vect x|\vect g, \vect z)$, is typically called the decoder.

Since the KL-divergence is non-negative, we obtain a lower bound on the data log-likelihood $\ln p_\theta(\vect x|\vect g)$ when considering the right hand side of the identity given in Eq.~\ref{eq:CVAE}. CVAEs minimize the negated version of this lower bound, \ie, 
\begin{equation}
\min_{\theta,\phi} D_{\operatorname{KL}}(q_\phi(\vect z|\vect x, \vect g), p(\vect z|\vect g)) - \frac{1}{N}\!\sum_{i=1}^N \ln p_\theta(\vect x|\vect g, \vect z^i),
\label{eq:conditional_vae}
\end{equation}
where the expectation $\mathbb{E}_{q_\phi(\vect z|\vect x, \vect g)}$ is approximated via $N$ samples $\vect z^i\sim q_\phi(\vect z|\vect x, \vect g)$. For simplicity of the exposition, we ignored the summation over the samples in the dataset $\cD$, and  provide the objective for training of a single pair $(\vect x, \vect g)$.

We next discuss how we combine those ingredients for diverse, controllable yet structurally coherent colorization.

\section{Consistency and Controllability for Colorization}
\begin{figure*}[t]
\centering

\includegraphics[width=\textwidth]{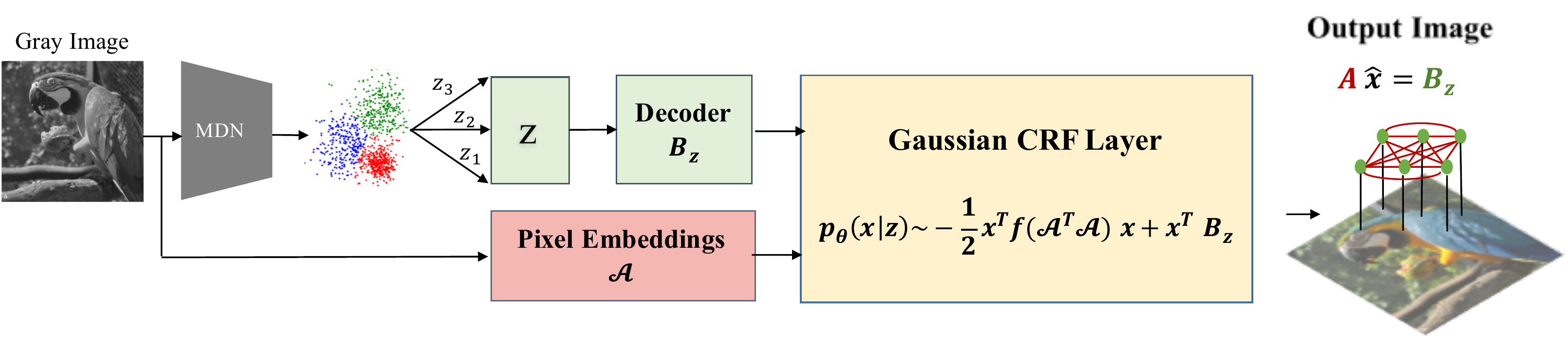}\\

\caption{A fully connected Gaussian Conditional Random Field (G-CRF) based VAE for diverse and globally coherent colorization. To generate diverse colorizations, we use a Mixture Density Network (MDN) to represent the multi-modal distribution of the color field embedding $\vect{z}$ given the gray-level image $\vect{g}$. At test  time, we sample multiple embeddings that are subsequently decoded to generate different colorizations. To ensure global consistency, we model the output space distribution of the decoder using a G-CRF.}
\label{fig:overview}

\end{figure*}
Our proposed colorization model has several appealing properties: (1) \textit{diversity}, \ie, it generates diverse and realistic colorizations for a single gray-level image; (2) \textit{global coherence}, enforced by explicitly modeling the output-space distribution of the generated color field using a fully connected Gaussian Conditional Random field (G-CRF); 
(3) \textit{controllability}, \ie, our model can consider external constraints at run time efficiently. For example, the user can enforce a given object to have a specific color or two separated regions to have the same colorization.

\afterpage{
\begin{figure*}[t!]
\centering   
\includegraphics[width=\linewidth ]{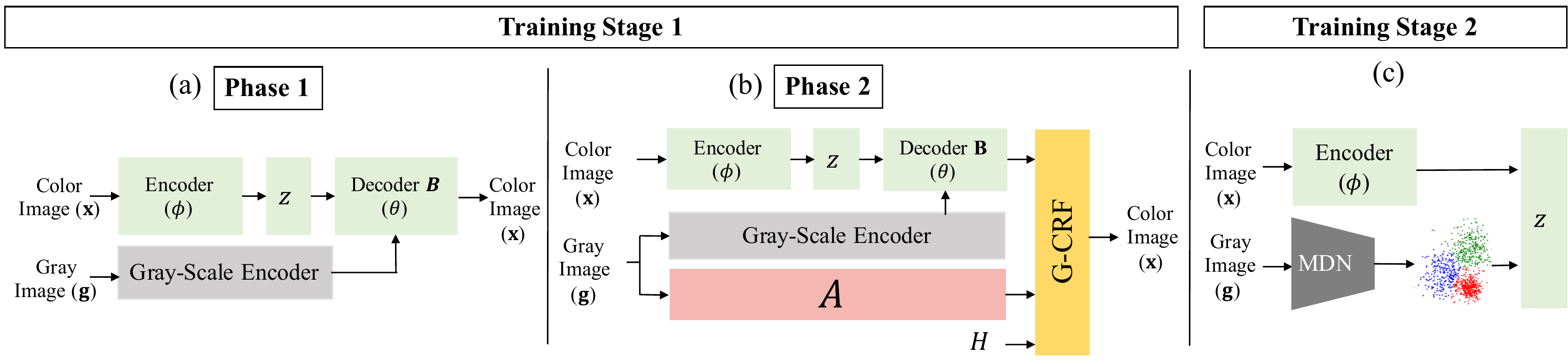}\\
\caption{Overview of the model architecture and the training procedure. In the first training stage, we learn a low dimensional embedding $\vect{z}$ of the \emph{color field} $\vect{x}$ conditioned on the gray-level image $\vect{g}$ using a VAE. To disentangle color from structure, we first learn the unary term $\vect{B}$ in \emph{phase 1}, then in \emph{phase 2}, learn a precision matrix that encodes the \emph{structure} of the image by imposing the constraint that pixels with similar intensities should have similar colorizations. To enable controllability, we use a training schedule specified in the matrix $\vect{H}$ to incrementally mask the decoded pixel colors in the unary term $\vect{B}$ and hence gradually rely on the $\vect{A}$ matrix to restore the colorization from the  unary term. In the second training stage, we use an MDN to learn a multi-modal distribution of the latent embedding given the gray-level image. }
\label{fig:architecture}

\end{figure*}
}
\subsection{Overview}
We provide an overview of our approach in  Fig.~\ref{fig:overview}. Given a gray-level image $\vect g$ with $P$ pixels, our goal is to produce different color fields $\vect{x} \in \mathbb{R}^{2P} $ consisting of two channels $\vect x_{a}\in\mathbb{R}^P$ and $\vect x_{b}\in\mathbb{R}^P$ in the \textit{Lab} color space. In addition, we  enforce  spatial coherence at a global scale and enable controllability using a Gaussian Markov random field which models the output space distribution.

To produce a diverse colorization, we  want to learn a multi-modal conditional distribution $p(\vect{x}|\vect{g})$ of the color field $\vect{x}$ given the gray-level image $\vect{g}$. However, learning this conditional is challenging since the color field $\vect{x}$ and the intensity field $\vect{g}$ are high dimensional. Hence, training samples for learning $p(\vect{x}|\vect{g})$ are sparsely scattered and the distribution is difficult to capture, even when using large datasets. Therefore, we assume the manifold hypothesis to hold, and we choose to learn a conditional $p(\vect x|\vect z, \vect g)$ based on low-dimensional embeddings $\vect z$ captured from $\vect{x}$ and $\vect{g}$, by using a variational auto-encoder which approximates the intractable posterior $p(\vect{z}|\vect{x},\vect{g})$ via an encoder. Deshpande \etal~\cite{deshpande2016} demonstrated that sampling from the approximation of the posterior results in low variance of the generated images. Following~\cite{deshpande2016}, we opt for a multi-stage training procedure to directly sample from $p(\vect{z}|\vect{g})$ as follows.

To capture the low-dimensional embedding, in a \emph{first training stage}, we use a variational auto-encoder to learn a parametric uni-modal Gaussian encoder distribution $q_{\phi}(\vect{z}|\vect{x},\vect{g})\sim \mathcal{N}(\vect \mu_{\phi}, \sigma_{\phi}^{2} \vect I )$ of the color field embedding $\vect{z}$ given both the gray-level image $\vect{g}$ and the color image $\vect{x}$ (Fig.~\ref{fig:architecture}~(a)). At the same time, we learn the parameters $\theta$ of the decoder $p_\theta(\vect x|\vect z, \vect g)$. 

Importantly, we note that the encoder $q_{\phi}(\vect{z}|\vect{x},\vect{g})$ takes advantage of both the color image $\vect x$ and the gray-level intensities $\vect g$ when mapping to the latent representation $\vect z$. Due to the use of the color image, we expect that this mapping can be captured to a reasonable degree using a uni-modal distribution, \ie, we use a Gaussian. 

However, multiple colorizations can be obtained from a gray-scale image $\vect g$ during inference. Hence, following Deshpande \etal~\cite{deshpande2016}, we don't expect a uni-modal distribution $p(\vect z|\vect g)$ to be accurate during testing, when only conditioning on the gray-level image $\vect g$.
 
To address this issue, in a \emph{second training stage}, we train a Mixture Density Network (MDN) $p_{\psi}(\vect{z}|\vect{g})$ to maximize the log-likelihood of embeddings $\vect z$ sampled from $q_{\phi}(\vect{z}|\vect{x},\vect{g})$ 
 (Fig.~\ref{fig:architecture}~(b)). 
 Intuitively, for a gray-level image, the MDN predicts the parameters of $M$ Gaussian components  each corresponding to a different colorization. The embedding $\vect{z}$ that was learned in the first stage is then tied to one of these components. The remaining components are optimized by close-by gray-level image embeddings.

At test time, $N$ different embeddings $\{\vect{z}\}_{k=1}^{N}$ are sampled from the MDN $p_{\psi}(\vect{z}| \vect{g})$ and transformed by the decoder into diverse colorizations, as we show in Fig.~\ref{fig:overview}. 

To encourage globally coherent colorizations and to ensure controllability, we use a fully connected G-CRF layer to model the output space distribution. The negative log-posterior of the G-CRF  has the form of a quadratic energy function:
\begin{equation}
 E( \vect x)= \frac{1}{2} \vect x^{T} \vect{A_{\vect g}} \vect x - \vect B_{\vect z, \vect g} \vect x.
\label{eq:gcrf}
\end{equation}
 It captures unary and higher order correlations (HOCs) between the pixels' colors for the \textit{a} and \textit{b} channels. Intuitively, the joint G-CRF enables the model to capture more global image statistics which turn out to yield more spatially coherent colorizations as we will show in Sec.~\ref{sec:exp}. The unary term $\vect B_{\vect z,\vect g}$ is obtained from the VAE decoder and encodes the color per pixel. The HOC term $\vect A_{\vect g} = f(\vect{\mathcal{A}}_{\vect g}^T\vect{\mathcal{A}}_{\vect g}$) is responsible for encoding the structure of the input image. It is a function of the inner product of low rank pixel embeddings $\vect{\mathcal{A}}_{\vect g}$, learned from the gray-level image and measuring the pairwise similarity between the pixels' intensities. The intuition is that pixels with similar intensities should have similar colorizations. The HOC term is shared between the different colorizations obtained at test time. Beyond global consistency, it also enables controllability by propagating user edits encoded in the unary term properly. Due to the symmetry of the HOC term, the quadratic energy function has a unique global minimum that can be obtained by solving the system of linear equations:
 \begin{equation}
\vect A_{\vect g} \vect x = \vect B_{\vect z,\vect g}.
\end{equation}
Subsequently, we  drop the dependency of $\vect A$ and $\vect B$ on $\vect g$ and $\vect z$ for notational simplicity.  
 
We now discuss how to perform inference in our model and how to learn the model parameters such that colorization and structure are disentangled and controllability is enabled by propagating  user strokes. 

\subsection{Inference}
\label{sec:Inference}
In order to ensure a globally consistent colorization, we take advantage of the structure in the image. To this end, we encourage two pixels to have similar colors if their intensities are similar. Thus, we want to minimize the difference between the color field $\vect x$ for the $a$ and $b$ channels and the weighted average of the colors at  similar pixels. More formally, we want to encourage the equalities  $\vect x_{a} = \vect{\hat S} \vect x_{a} $ and  $\vect x_{b} = \vect{\hat S} \vect x_{b} $, where  $\vect{\hat S} = \operatorname{softmax}(\vect{\mathcal{A}^{T}} \vect{ \mathcal{A}}) $ is a similarity matrix obtained from applying a softmax function to every row of the matrix resulting from $\vect{\mathcal{A}^{T}} \vect{ \mathcal{A}}$. To simplify, we use the block-structured matrix $\vect S = \operatorname{diag}(\vect{\hat S}, \vect{\hat S})$.

In addition to capturing the structure, we obtain the color prior and  controllability by encoding the user input in the computed unary term $\vect B$. Hence, we add the constraint $\vect H \vect x = \vect \alpha$, where $\vect H$ is a diagonal matrix with 0 and 1 entries corresponding to whether the pixel's value isn't or is specified by the user, and $\vect \alpha$ a vector encoding the color each pixel should be set to. 

With the aforementioned intuition at hand we obtain the quadratic energy function to be minimized as: 
$$  E_{\theta,\vect g,\vect z}(\vect{x})=\frac{1}{2}  \|(\vect I - \vect S) \vect x\|^{2} + \frac{1}{2}\beta \|\vect H \vect x -\vect \alpha\|^{2}, $$ with $\beta$ being a hyper-parameter. This corresponds to a quadratic energy function of the form $ \frac{1}{2} \vect x^{T} \vect{A} \vect x + \vect B \vect x  + C$, where $\vect A= (\vect S-\vect I )^{T}  (\vect S-\vect I ) + \beta \vect H^{T} \vect H $, $\vect B= -2  \beta \vect \alpha^{T} \vect H $ and $C = \beta \vect \alpha^{T} \vect \alpha $. It's immediately apparent that the unary term only encodes  color statistics while the HOC term is only responsible for  structural consistency. Intuitively, the conditional $p_{\theta}(\vect{x}|\vect{g},\vect{z})$ is interpreted as a Gaussian multi-variate density:
\begin{equation}
p_{\theta}(\vect{x}|\vect{z},\vect g)\propto \exp(-E_{\theta,\vect g,\vect z}(\vect{x}) ) ,
\end{equation}
parametrized by the above defined energy function $E_{\theta, \vect g, \vect z}$. It can be easily checked that $\vect A$ is a positive definite full rank matrix. Hence, for a strictly positive definite matrix, inference is reduced to solving a linear system of equations: 
\begin{equation}
((\vect I - \vect S)^T (\vect I - \vect S) +\beta \vect H^{T} \vect H ) \vect x= \beta \vect H^{T} \vect \alpha.
\label{eq:linear_system}
\end{equation}
We solve the linear system above using the LU decomposition of the $\vect A$ matrix. How to learn the terms $\vect \alpha$ and $\vect S$ will be explained in the following. 

\begin{figure*}[!t]
\centering
\includegraphics[width=0.8\linewidth]{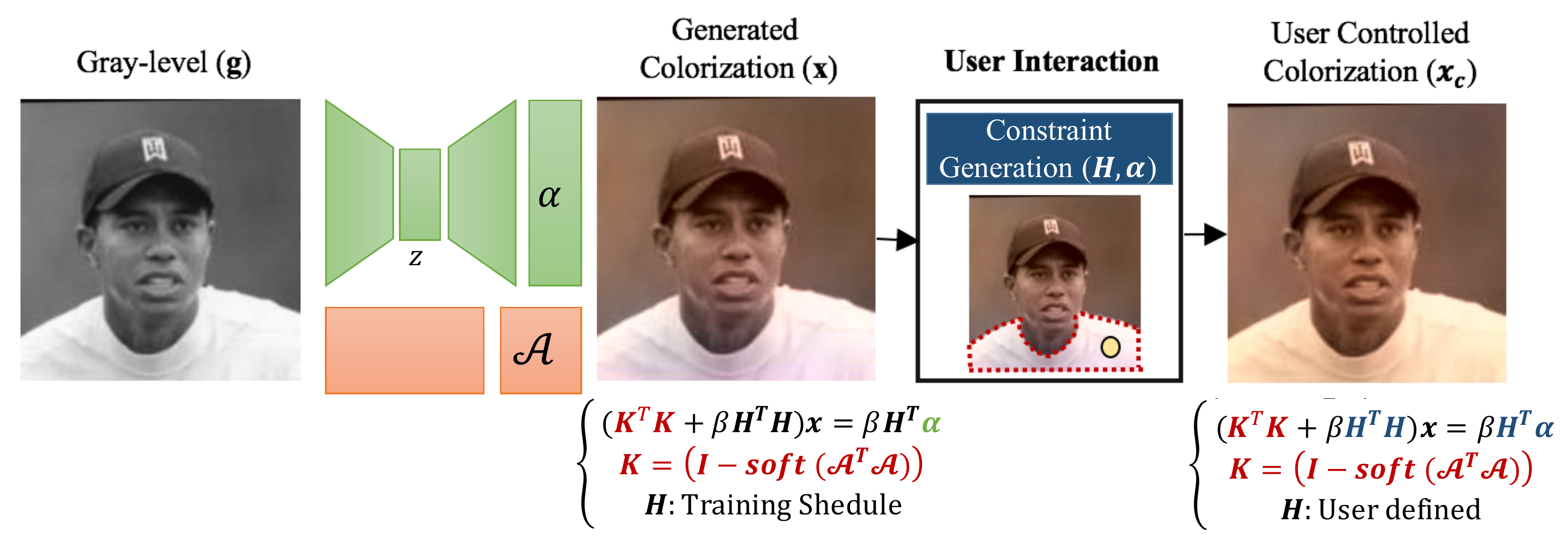}
\caption{Controllability: Given a gray-level image, we learn to disentangle structure from colorization. The HOC term is used to propagate sparse user edits encoded in the $\vect H$ and $\vect \alpha$ terms. }

\label{fig:controlled}
\end{figure*}

\subsection{Learning}
\label{sec:Learning}
We now present the two training stages illustrated in Fig.~\ref{fig:architecture} to ensure color and structure disentanglement and to produce diverse colorizations. We also discuss the modifications to the loss given in Eq.~\ref{eq:conditional_vae} during each stage. 

\noindent{\bf Stage 1: Training a Structured Output Space Variational Auto-Encoder:} 
During the first training stage, we use the variational auto-encoder formulation to learn a low-dimensional embedding for a given color field. This stage is divided into two phases to ensure color and structure disentanglement. In a first phase, we learn the unary term produced by the VAE decoder. In the second phase, we fix the weights of the VAE apart from the  decoder's two top-most layers and learn a $D$-dimensional embedding matrix $\vect{\mathcal{A}} \in \mathbb{R}^{D \times P }$ for the $P$ pixels from the gray-level image. The matrix $\vect{\hat S}$ obtained from applying a softmax to every row of $\vect{\mathcal{A}}^{T}  \vect{\mathcal{A}} $ is used to encourage a smoothness prior $\vect x = \vect{ S} \vect x$ for the $a$ and $b$ channels. In order to ensure that the $\vect{ S}$ matrix  learns  the structure required for the controllability stage, where sparse user edits need to be propagated, we follow a training schedule where the unary terms are masked gradually using the $\vect H$ matrix. The input image is reconstructed from the sparse unary entries using the learned structure. When colorization from sparse user edits is desired, we solve the linear system from Eq.~\ref{eq:linear_system} for the learned HOC term and an $\vect H$ matrix and $\vect \alpha$ term encoding the user edits, as illustrated in Fig.~\ref{fig:controlled}. We explain the details of the training schedule in the experimental section.

Given the new formulation of the G-CRF posterior, the program for the first training stage reads as follows: 
\bea
\hspace{-0.2cm}\underset{\phi, \theta}{\min}  \hspace{0.1cm} 
D_{\operatorname{KL}}(\mathcal{N}( \vect{\mu}_{\phi}  , \vect{\sigma}_{\phi}^{2} \vect{I})  ),\mathcal{N}( 0 ,\vect{I})  )\! -\!\frac{1}{N}\! \sum_{i=1}^{N}\!   \ln p_{\theta}(\vect{x}|\vect{z}^{(i)}\!\!,\!\vect{g}) 
 \hspace{0.1cm} \text{s.t.} \vect{z}^{(i)}\!\sim\! \mathcal{N}( \vect{\mu}_{\phi}, \vect{\sigma}_{\phi}^{2} \vect{I}).
 \label{eq:obj}
\eea
Subsequently we use the term $L$ to refer to the objective function of this program. 

\noindent{\bf Stage 2: Training a Mixture Density Network (MDN):}  
Since a color image $\vect x$ is not available during testing, in the second training stage, we capture the approximate posterior $q_\phi(\vect{z}|\vect{x},\vect{g})$, a Gaussian which was learned in the first training stage, using a parametric distribution $p_{\psi}(\vect z|\vect g)$.
%
Due to the dependence on the color image $\vect x$ we expect the approximate posterior $q_\phi(\vect{z}|\vect{x},\vect{g})$ to be easier to model than $p_{\psi}(\vect z|\vect g)$. 
Therefore, we let $p_{\psi}(\vect z|\vect g)$ be a Gaussian Mixture Model (GMM) with $M$ components. 
Its means, variances, and component weights are parameterized via a mixture density network (MDN) with parameters $\psi$. 
Intuitively, for a given gray-level image, we expect the $M$  components to correspond to different colorizations. The colorfield embedding $\vect{z}$ learned from the first training stage is mapped to one of the components by minimizing the negative conditional log-likelihood, \ie, by minimizing: 
\begin{equation}
-\ln p_{\psi}(\vect z|\vect g)= -\ln \sum_{{i=1}}^{M} \pi^{(i)}_{\vect g,\psi} \mathcal{N}(\vect z|\vect \mu^{(i)}_{ g,\psi},\sigma).
\label{eq:gmm1}
\end{equation}
Hereby, $\pi_{\vect g,\psi}^{(i)}$, $\vect \mu_{\vect g,\psi}^{(i)}$ and $\sigma$ refer to, respectively, the mixture coefficients, the means and a fixed co-variance of the GMM learned by an MDN network parametrized by $\psi$.
However, minimizing $-\ln p_{\psi}(\vect z|\vect g)$ is hard as it involves the computation of the logarithm of a summation over the different exponential components. To avoid this, we explicitly assign the code $\vect z$ to that Gaussian component $m$, which has its mean closest to $\vect z$, \ie, $m= \underset{i}{\mathrm{argmin}}  \| \vect z - \vect \mu_{g,\psi}^{(i)} \| $. Hence, the negative log-likelihood loss $-\ln p_{\psi}(\vect z|\vect g)$ is reduced to solving the following program: 
\bea
\underset{\psi}{\min}  \hspace{-0.0cm} 
-\ln \pi_{\vect g, \psi}^{(m)}+ \frac{ \| \vect z - \vect \mu_{\vect g, \psi}^{(m)}  \|^{2}  }{2 \sigma^{2}}   &&
\text{s.t. } 
\hspace{0.1cm}\left\{\begin{array}{l}
\vect{z} \sim q_\phi(\vect{z}|\vect{x},\vect{g}) = \mathcal{N}( \vect{\mu}_{\phi}  , \vect{\sigma}_{\phi}^{2} \vect{I})   \\
m= \underset{i\in\{1, \ldots, M\}}{\mathrm{argmin}} \| \vect z - \vect \mu_{g,\psi}^{(i)} \|
\end{array}\right..\label{eq:objMDN1}
\eea
Note that the latent samples $\vect z$ are obtained from the approximate posterior $q_\phi(\vect z|\vect x, \vect g)$ learned in the first stage.

\begin{figure*}[t]
    \centering 

\includegraphics[width=4.8in, height=3.6in]{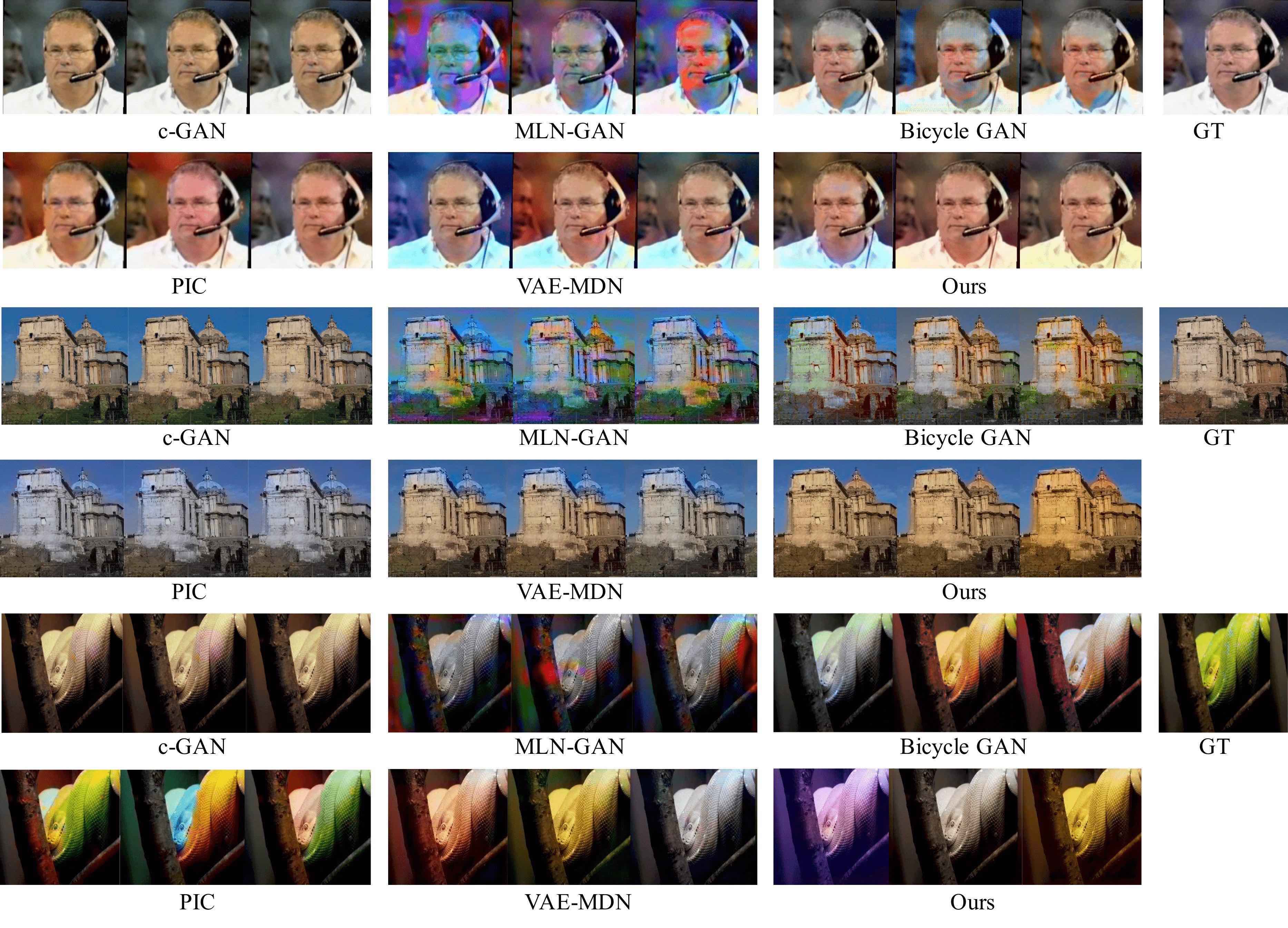} 
 \label{fig:rex_ritter}
\vspace{-0.8cm}
  \caption{Qualitative comparison of diverse colorizations obtained from c-GAN~\cite{isola2017}, MLN-GAN~\cite{Cao2017}, BicycleGAN~\cite{zhu2017toward}, PIC~\cite{royer2017}, VAE-MDN~\cite{deshpande2016} and our approach. }
\label{fig:baseline_comparison}

\end{figure*}


\section{Results}
\label{sec:exp}
Next, we present quantitative and qualitative results on three datasets of increasing color field complexity: (1) the Labelled Faces in the Wild dataset (LFW)~\cite{learned2016labeled}, which consists of 13,233 face images aligned by deep funneling~\cite{huang2012learning}; (2) the LSUN-Church dataset~\cite{yu2015lsun} containing 126,227 images and (3) the validation set of ILSVRC-2015
(ImageNet-Val)~\cite{russakovsky2015imagenet} with 50,000 images.
We  compare the diverse colorizations obtained by our model with three baselines representing three different generative models: (1) the Conditional Generative Adversarial Network~\cite{isola2017,Cao2017,zhu2017toward}; (2) the Variational Auto-encoder with MDN~\cite{deshpande2016}; and (3) the Probabilistic Image Colorization model~\cite{royer2017} based on PixelCNN. Note that~\cite{deshpande2016} presents a comparison between VAE-MDN and a conditional VAE, demonstrating the benefits of the VAE-MDN approach. 

\subsection{Baselines}
\label{sec:baselines}
\begin{table}[!t]
\centering
\caption{ \small{Results of the user study \scriptsize(\% of the model in \textbf{bold} winning).}}
\setlength\tabcolsep{3pt}
\begin{tabular}{cccc} 
\toprule
&\emph{\scriptsize \textbf{Ours} vs VAE-MDN}  & \emph{\scriptsize \textbf{Ours} vs PIC} & \emph{\scriptsize \textbf{VAE-MDN} vs PIC} \\ \midrule
\emph{\scriptsize LFW} & \scriptsize 61.12 \% & \scriptsize 59.04 \%  & \scriptsize 57.17 \%   \\
\emph{\scriptsize LSUN-Church} & \scriptsize 66.89 \%  &\scriptsize 71.61 \%  & \scriptsize 54.46 \%   \\
\emph{\scriptsize ILSVRC-2015} & \scriptsize 54.79 \%  & \scriptsize 66.98 \%  & \scriptsize 62.88\%   \\
 \bottomrule
\label{tab:user_study}
\end{tabular}
\end{table}

\begin{table}[!t]
\centering
\caption{Quantitaive comparison with  baselines. We use the error-of-best per pixel (Eob.), the variance (Var.), the mean structural similarity SSIM across all pairs of colorizations generated for one image (SSIM.)
and the training time (Train.) as performance metrics.}
\setlength\tabcolsep{0pt}
\begin{tabular}{cccccccccccccccccc} \toprule
\multirow{4}{*}{{\scriptsize Method}} & \multicolumn{4}{c}{{\scriptsize LFW}} & \multicolumn{4}{c}{{\scriptsize LSUN-Church}} & \multicolumn{4}{c}{\scriptsize{\scriptsize ILSVRC-2015}} \\
\cmidrule(lr){2-5} \cmidrule(lr){6-9} \cmidrule(lr){10-13} 

& {\scriptsize eob.}  &  {\scriptsize Var.}  &   {\scriptsize SSIM.} &   {\scriptsize Train.}   
& {\scriptsize eob.}&  {\scriptsize Var.}   &  {\scriptsize SSIM.} &   {\scriptsize Train.} 
& {\scriptsize eob.} & {\scriptsize Var.}  &  {\scriptsize SSIM.} &   {\scriptsize Train.}   \\

{\scriptsize  c-GAN\cite{isola2017} }  &  {\scriptsize$.047$} &  {\scriptsize $8.40 e^{-6}$} &   {\scriptsize $.92$}  & {\scriptsize$\sim\!\!\mathbf{4}$h }&
 {\scriptsize$.048$} & {\scriptsize $6.20 e^{-6}$}   &  {\scriptsize$.94$}    & {\scriptsize$\sim\!\!\mathbf{39}$h } &  
 {\scriptsize$.048$} & {\scriptsize $8.88 e^{-6}$ }    &{\scriptsize $.91$}  & {\scriptsize$\sim\!\!\mathbf{18}$h}    \\

{\scriptsize  MLN-GAN\cite{Cao2017}} & {\scriptsize$.057$}   &   {\scriptsize $2.83 e^{-2} $} &   {\scriptsize $\mathbf{.12}$}  & {\scriptsize$\sim\!\!\mathbf{4}$h }&  
{\scriptsize$.051$} & {\scriptsize $2.48e^{-2}$}   &  {\scriptsize$\mathbf{.34}$}    & {\scriptsize$\sim\!\!\mathbf{39}$h } & 
{\scriptsize$.063$}  & {\scriptsize $1.73 e^{-2}$ }    &{\scriptsize $.38$}  & {\scriptsize$\sim\!\!\mathbf{18}$h}    \\

{\scriptsize  BicycleGAN\cite{zhu2017toward}}&{\scriptsize $.045$}   &   {\scriptsize $6.50 e^{-3}$} &   {\scriptsize $.51$}  & {\scriptsize$\sim\!\!\mathbf{4}$h } &
{\scriptsize $.048$}   & {\scriptsize $2.20 e^{-2}$}   &  {\scriptsize$.38$}    & {\scriptsize$\sim\!\!\mathbf{39}$h } &
{\scriptsize$.042$}  &  {\scriptsize $2.20 e^{-2}$ }    &{\scriptsize $\mathbf{.15}$}  & {\scriptsize$\sim\!\!\mathbf{18}$h}    \\

{\scriptsize VAE-MDN\cite{deshpande2016}}  &   {\scriptsize$.035$}   &   {\scriptsize $1.81 e^{-2}$}   &  {\scriptsize $.49$}  & {\scriptsize $\sim\!\!\mathbf{4}$h } &   {\scriptsize$.028$}   &   {\scriptsize $\mathbf{1.05e^{-2}}$ }     &  {\scriptsize $ .77$}    & {\scriptsize $\sim\!\!\mathbf{39}$h }&  
  {\scriptsize$.033$}  &   {\scriptsize $7.17e^{-3}$ }    & {\scriptsize $.48$}  & {\scriptsize $\sim\!\!\mathbf{18}$h }   \\ 

{\scriptsize PIC\cite{royer2017}} &  {\scriptsize $.043$}   &   {\scriptsize  $\mathbf{5.32 e^{-2}}$}   &   {\scriptsize $.36$}  &  {\scriptsize $\sim\!\!48$h} &  {\scriptsize $.047$} &    {\scriptsize $7.40 e^{-5}$}  &   {\scriptsize $.91$}    &  {\scriptsize $\sim\!\!144$h} & 
{\scriptsize$.035$} & {\scriptsize  $\mathbf{6.74 e^{-2}}$}    & {\scriptsize  $.19$ }  &  {\scriptsize  $\sim\!\!96$h}    \\ 

{\scriptsize \bf Ours}&   {\scriptsize$ \mathbf{11 e^{-5}}$} &   {\scriptsize $8.86 e^{-3} $}  &   {\scriptsize $.61$} &  {\scriptsize $\sim\!\!\mathbf{4}$h} &
{\scriptsize$ \mathbf{93 e^{-6}}$}& {\scriptsize $ 1.17 e^{-2}$ } &  {\scriptsize $ .83$}  & {\scriptsize $\sim\!\!\mathbf{39}$h }&  
{\scriptsize $\mathbf{12e^{-5}}$} &   {\scriptsize $8.80 e^{-3}$}   & {\scriptsize $.52 $} & {\scriptsize $\sim\!\!\mathbf{18}$h } \\ \bottomrule
\label{tab:quantitative_results}

\end{tabular}
\end{table}

\noindent{\bf Conditional Generative Adversarial Network:} We compare our approach with three GAN models: the c-GAN architecture proposed by Isola \etal~\cite{isola2017}, the GAN with multi-layer noise by Cao \etal~\cite{Cao2017} and the BicycleGAN by Zhu \etal~\cite{zhu2017toward}. 

\noindent{\bf Variational Auto-Encoder with Mixture Density Network (VAE-MDN):} 
The architecture by Deshpande \etal~\cite{deshpande2016} trains an MDN based auto-encoder to generate different colorizations. It is the basis for our method. 

\noindent{\bf Probabilistic Image Colorization (PIC):} 
The PIC model proposed by Royer \etal~\cite{royer2017} uses a CNN network to learn an embedding of a gray-level images, which is then used as input for a PixelCNN network. 

\noindent{\bf Comparison with Baselines: }
We qualitatively compare the diversity and global spatial  consistency of the colorizations obtained by our models with the ones generated by the aforementioned baselines, in Figures \ref{fig:first_fig} and \ref{fig:baseline_comparison}. We observe that our approach is the only one which generates a consistent colorization of the skin of the girl in Fig.~\ref{fig:first_fig}. We are also able to uniformly color the ground, the snake, and the actor's coat in Fig.~\ref{fig:baseline_comparison}.

For global consistency evaluation, we perform a user study, presented in Tab.~\ref{tab:user_study}, where participants are asked to select the more realistic image from a pair of images at a time. We restrict the study to the three approaches with the overall lowest error-of-best (eob) per pixel  reported in Tab.~\ref{tab:quantitative_results}, namely VAE-MDN, PIC and our model. We use the clicking speed to filter out inattentive participants. Participants did neither know the paper content nor were the methods revealed to them. We gathered 5,374 votes from 271 unique users. The results show that users prefer results obtained with the proposed approach.

To evaluate diversity, we use two metrics: (1) the variance of diverse colorizations and (2) the mean structural similarity \emph{SSIM}~\cite{wang2004image} across all pairs of colorizations generated for one image. We report our results in Tab.~\ref{tab:quantitative_results}. 

\begin{figure}[!t]
\centering

\includegraphics[width=4.in, height=3.5in]{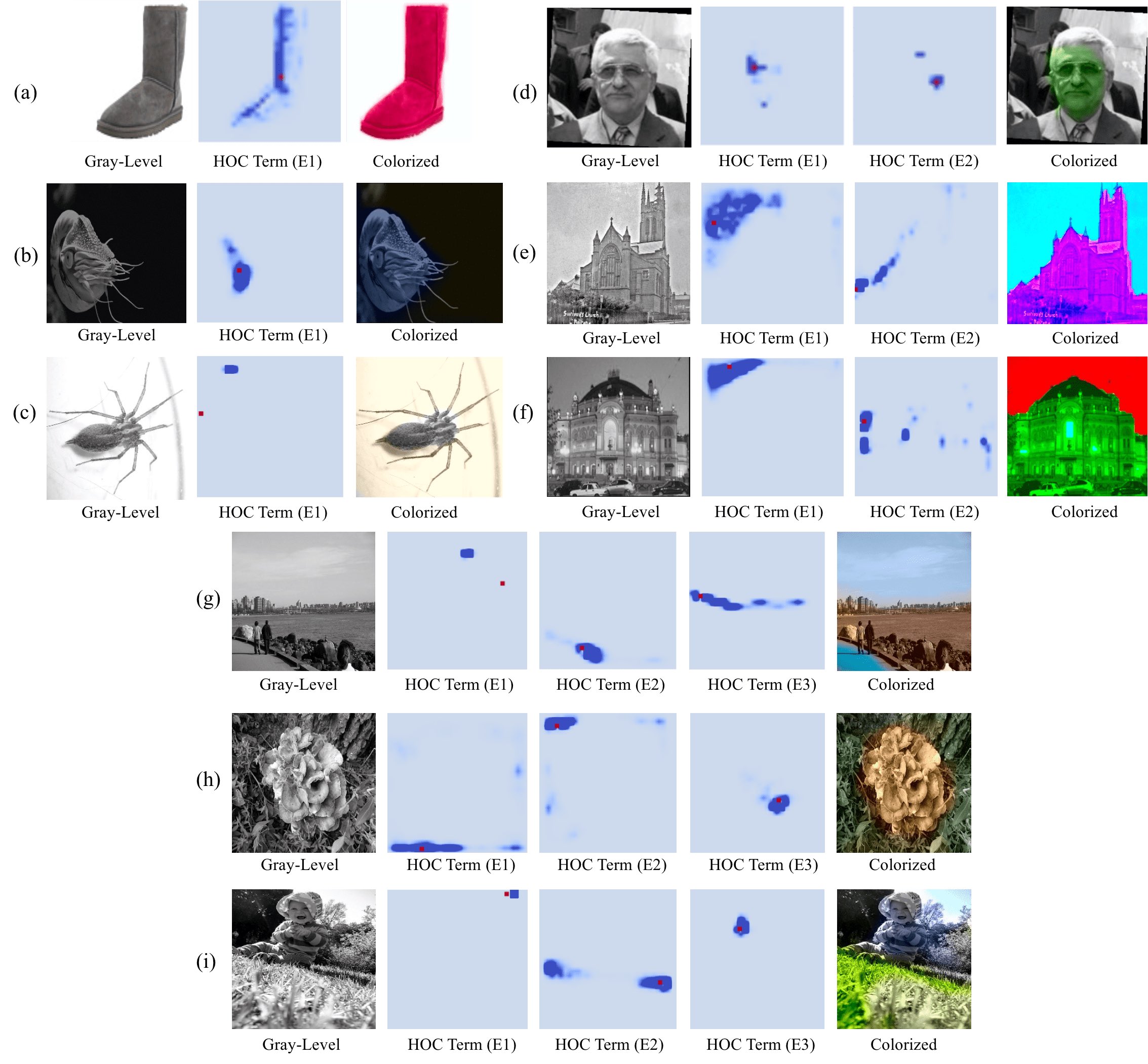}\\

\caption{ Controllability: Colorization from sparse user edits. }
\label{fig:controllability}
\end{figure}

\noindent{\bf Global Consistency:} Our model noticeably outperforms all the baselines in producing spatially coherent results as demonstrated by the user study. PIC generates very diversified samples for the LFW and ILSVRC-2015 datasets but lacks long range spatial dependencies because of the auto-regressive nature of the model. For example, the snake in the second row of Fig.~\ref{fig:baseline_comparison} has different colors for the head and the tail, and the woman's skin tone is inconsistent in Fig.~\ref{fig:first_fig}. The VAE-MDN, BicycleGAN and MLN-GAN outputs are sometimes speckled and objects are not uniformly colored. For example, parts of the dome of the building in the second row of Fig.~\ref{fig:baseline_comparison} are confused to be part of the sky and the shirt in the third row is speckled. In contrast, our model is capable of capturing complex long range dependencies. This is confirmed by the user study. 

\noindent{\bf Diversity:} Across all  datasets, c-GAN suffers from mode collapse and is  frequently unable to produce diverse colorizations. 
The PIC, MLN-GAN and BicycleGAN models yield the most diverse results at the expense of photo-realism. Our model produces diverse results while ensuring long range spatial consistency.

\begin{table}[!t]

\centering

{\small
\centering
\caption{\footnotesize{Average PSNR (dB) (higher is better) vs.\ number of revealed points ($|\vect{H}|$).}}
\label{tab:controllability_result}
\setlength\tabcolsep{3pt}
\begin{tabular}{c|ccc|ccc|ccc|ccc|ccc} \toprule

 & \multicolumn{3}{c|}{\scriptsize Levin et al.~\cite{levin2004}} & \multicolumn{3}{c|}{\scriptsize Endo et al.~\cite{endoEG2016}} & \multicolumn{3}{c|}{\scriptsize Barron et al.~\cite{barron2016fast}} & \multicolumn{3}{c|}{\scriptsize Zhang et al.~\cite{zhang2017}} & \multicolumn{3}{c}{\scriptsize  \textbf{Ours}} \\ \midrule
\emph{$|H|$} & \scriptsize 10 & \scriptsize 50   & \scriptsize 100
& \scriptsize 10 & \scriptsize 50   & \scriptsize 100
& \scriptsize 10 & \scriptsize 50  & \scriptsize 100
& \scriptsize 10 & \scriptsize 50  & \scriptsize 100
& \scriptsize 10 & \scriptsize 50  & \scriptsize 100
\\ \midrule

\scriptsize \textbf{PSNR}  & \scriptsize 26.5 & \scriptsize 28.5 &  \scriptsize 30
& \scriptsize 24.8 & \scriptsize 25.9 & \scriptsize 26
& \scriptsize 25.3 & \scriptsize 28 & \scriptsize 29
& \scriptsize 28 & \scriptsize 30.2 & \scriptsize 31.5
& \scriptsize 26.7 & \scriptsize 29.3 & \scriptsize 30.4 \\
\bottomrule

\end{tabular}
}

\end{table}

\noindent{\bf Controllability:} For the controllably experiments, we set the $\beta$ hyper-parameter to 1 during training and to 5 during testing. We opt for the following training schedule, to force the model to encode the structure required to propagate sparse user inputs in the controllability experiments: We train the unary branch for 15 epochs (Stage1, Phase1), then train the HOC term for 15 epochs as well (Stage1, Phase2). We use the diagonal matrix $\vect H$ to randomly specify $L$ pixels which colors are encoded by the unary branch  $\vect \alpha$. We decrease $L$ following a training schedule from $100\%$ to $75\%$, $50\%$, $25\%$ then $10\%$ of the total number of pixels after respectively epochs $2$, $4$, $6$, $8$, $10$ and $12$. Note that additional stages could be added to the training schedule to accommodate for complex datasets where very sparse user input is desired. 
In Fig.~\ref{fig:controllability}, we show that with a single pixel as a user edit ($E1$), we are able to colorize a boot in pink, a sea coral in blue and the background behind the spider in yellow in respectively Fig.~\ref{fig:controllability}~(a-c). With two edits ($E1$ and $E2$), we colorize a face in green  (Zhang \etal~\cite{zhang2017} use 3 edits) in Fig.~\ref{fig:controllability}~(d) and the sky and the building in different colors in Fig.~\ref{fig:controllability}~(e,f). With three user edits ($E1$, $E2$ and $E3$), we show that we can colorize more complex images in Fig.~\ref{fig:controllability}~(g-i). We show the edits $E$ using red markers. We visualize the attention weights per pixel, corresponding to the pixel's row in the similarity matrix $\vect S$, in blue, where darker shades correspond to stronger correlations.

 Quantitatively, we report the average PSNR for 10, 50 and 100 edits on the ImageNet test set in Tab.~\ref{tab:controllability_result}, where edits (points) corresponding to randomly selected $7 \times 7$ patches are revealed to the algorithm. We observe that our method achieves slightly better results than the one proposed by Levin \etal~\cite{levin2004} as our algorithms learns for every pixel color an `attention  mechanism' over all the pixels in the image while Levin \etal~impose local smoothness. 

\begin{figure*}[!t]
\centering

\includegraphics[width=4.8in, height=1.2in]{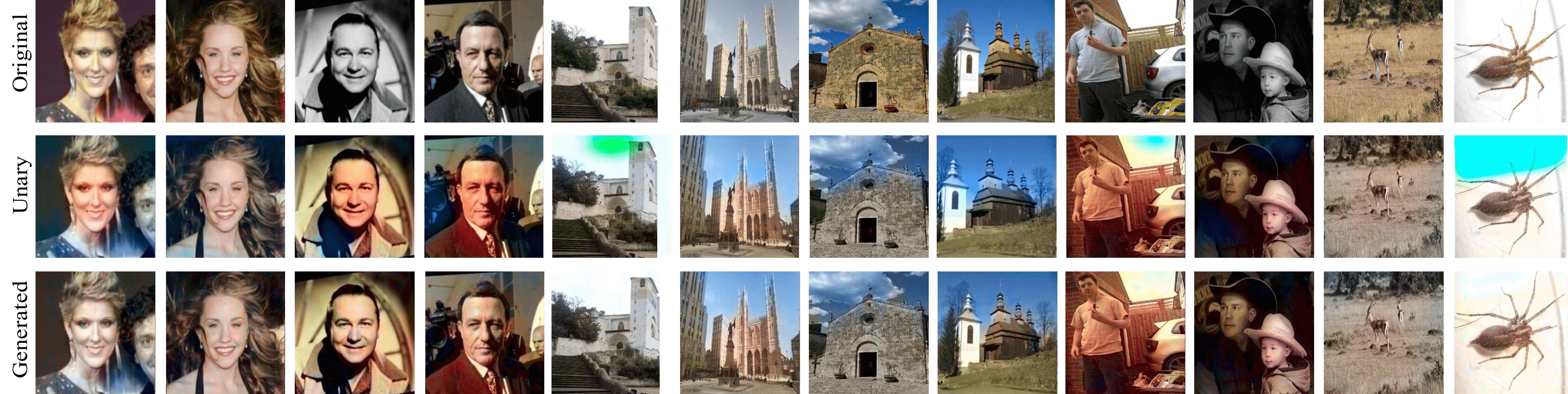}

\caption{ Visualization of the unary term. The first row corresponds to the ground truth image. We visualize one possible colorization in the third row and its corresponding unary term in the second row.}
\label{fig:unary_vis}

\end{figure*}

\begin{figure}[!t]
    \centering
    \includegraphics[width=0.9\textwidth]{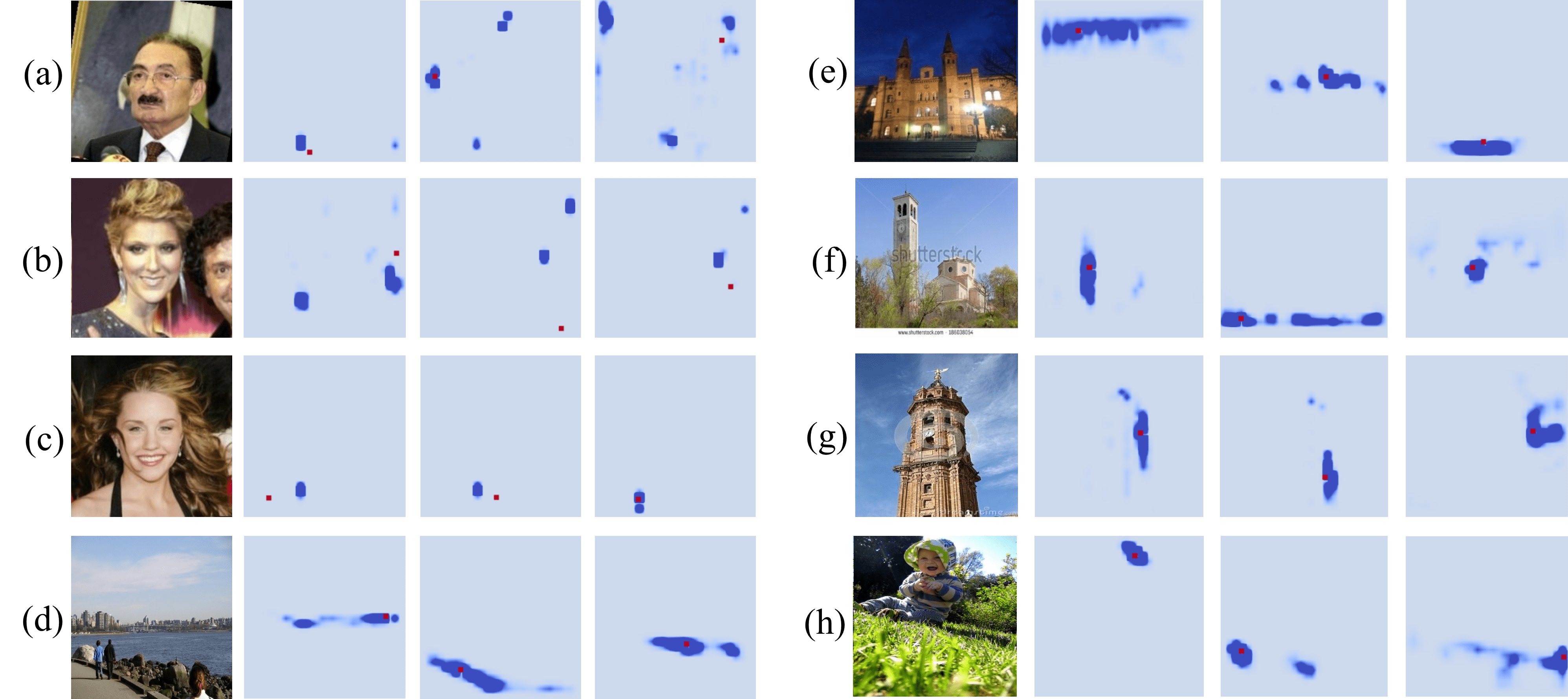}
   
    \caption{Visualization of the HOC term. For every example, we show the ground truth image and three HOC terms corresponding to three different pixels marked in red.}
    
 \label{fig:binary_vis}
 
\end{figure}
\noindent{\bf Visualization of the HOC and Unary Terms:} In order to obtain more insights into the model's dynamics, we visualize the unary terms, $\vect B$, and the HOC terms, $\vect A$, in respectively Fig.~\ref{fig:unary_vis} and Fig.~\ref{fig:binary_vis}. As  illustrated in Fig.~\ref{fig:binary_vis}, the HOC term has learned complex long range pixel affinities through end-to-end training. The results in Fig.~\ref{fig:unary_vis} further suggest that the unary term outputs a colorization with possibly some noise or inconsistencies that the HOC term fixes to ensure global coherency. For example, for the picture in the second column in Fig.~\ref{fig:unary_vis}, the colors of the face, chest and shoulder predicted by the unary term are not consistent, and were fixed by the binary term which captured the long range correlation as it is shown in Fig.~\ref{fig:binary_vis}~(c).

We notice different interesting strategies for encoding the long range correlations: On the LSUN-Church dataset, the model encourages local smoothness as every pixel seems to be strongly correlated to its neighbors. This is the case for the sky in Fig.~\ref{fig:binary_vis}~(e). The model trained on the LFW dataset, however encoded long range correlation. To ensure consistency over a large area, it chooses some reference pixels and correlates every pixel in the area, as  can be seen in Fig.~\ref{fig:binary_vis}~(c).

We provide more results and details of the employed deep net architectures in the supplementary material.


\section{Conclusion}

We proposed a Gaussian conditional random field based variational auto-encoder formulation for colorization and illustrated its efficacy on a variety of benchmark datasets, outperforming existing methods. The developed approach goes beyond existing methods in that it doesn't only model the ambiguity which is inherent to the colorization task, but also takes into account structural consistency.

\noindent\textbf{Acknowledgments:} This material is based upon work supported in part by the National Science Foundation under Grant No.~1718221, Samsung, and 3M. We thank NVIDIA for providing the GPUs used for this research.

\bibliographystyle{splncs}
\bibliography{egbib}
\clearpage



\title{Supplementary Material: Structural Consistency and Controllability for Diverse Colorization} 


\author{Safa Messaoud \and
David Forsyth\and
Alexander G. Schwing}
\institute{University of Illinois at Urbana-Champaign, USA}



\maketitle

\section{Introduction}
We use a variational auto-encoder (VAE) model enriched with a Mixture Density Network (MDN) and a Gaussian Conditional Random field (G-CRF) to generate diverse and globally consistent colorizations while enabling controllability through sparse user edits. In the following, we will derive closed form expressions of the derivative of the loss with respect to the G-CRF layer inputs, \ie, the unary term $\vect B$ and the HOC term $\vect A$ (Sec.~\ref{sec:learning}). We next present the model's architecture (Sec.~\ref{sec:architecture}) and additional results on the LFW, LSUN-Church, ILSVRC-2015 (Sec.~\ref{sec:add_results}) and ImageNet (Sec.~\ref{sec:ImageNetresults}) datasets. Finally, we explore endowing VAE and BEGAN \cite{berthelot2017began} with a structured output-space distribution through the G-CRF formulation for image generation (Sec.~\ref{sec:Beyond Colorization}).

\section{Learning - Gradients of the G-CRF parameters }
\label{sec:learning}
During the first training phase's forward pass, the G-CRF receives the unary term $\vect{B}$ and the HOC term $\vect{A}$, and outputs the reconstructed color field after solving the linear system given in Eq. 4 of the main paper. In the backward pass, the G-CRF layer receives the gradient of the objective function $L$ of the program given in Eq. 7 in the main paper with respect to $\vect{x}$ and computes closed form expressions for the gradient of the loss with respect to $\vect{A}$ and $\vect{B}$. Using the chain rule, the gradients of the remaining parameters can be expressed in terms of $\frac{\partial L}{\partial \vect{A} }$ and $\frac{\partial L}{\partial \vect{B}}$. 

Note that the gradient of the unary term decomposes as $\frac{\partial L}{\partial \vect x}=\frac{\partial L}{\partial \vect B} \frac{\partial \vect B}{\partial \vect x}$. By taking Eq. 4 in the main paper into account, we compute $\frac{\partial \vect B}{\partial \vect x} =\vect A$. Hence, it can be easily verified that a closed form expression of the gradient of the loss with respect to the unary term $\vect B$ corresponds to solving the following system of linear equations: 
\begin{equation}  \tag{10}
\vect A  \frac{\partial L}{\partial \vect{B} } = \frac{\partial L}{\partial \vect{x} }.
\label{eq:grad_B}
\end{equation}

Similarly, we derive the gradient of the pairwise term by application of the chain rule: $\frac{\partial L}{\partial \vect A_{i,j}}=\frac{\partial L}{\partial \vect B_{i}}\frac{\partial \vect B_{i}}{\partial \vect A_{i,j}}=\frac{\partial L}{\partial \vect B_{i}} \vect x_{j}$. Hence:
\begin{equation} \tag{11}
\frac{\partial L}{\partial \vect{A} } =  \frac{\partial L}{\partial \vect{B} } \cdot \vect{x}^{T}=\frac{\partial L}{\partial \vect{B} } \otimes \vect{x}.
\label{eq:grad_A}
\end{equation}

\section{Architecture and Implementation Details}
\label{sec:architecture}

In this section, we  discuss the architecture of the encoder, the MDN, the decoder, the structure encoder and the G-CRF layer. 
We introduce the following notation for describing the different architectures. We denote by $C_{a}(k,s,n)$ a convolutional layer with kernel size $k$, stride $s$, $n$ output channels and activation $a$. We write $B$ for a batch normalization layer, $U(f)$ for bilinear up-sampling with scale factor $f$ and $FC(n)$ for fully connected layer with $n$ output channels. To regularize, we use dropout for fully connected layers. 

\noindent{\bf Encoder:} 
The \textit{encoder} network learns an approximate posterior $q_\phi(\vect z|\vect x, \vect g)$ conditioned on the gray-level image representation and the color image. Its input is a colorfield of size $64 \times 64 \times 2$ and it generates an embedding $z$ of size $64$. The architecture is  as follows: $64 \times 64 \times 2 \rightarrow C_{ReLU}(5,2,128) \rightarrow B \rightarrow C_{ReLU}(5,2,256)\rightarrow B \rightarrow C_{ReLU}(5,2,512) \rightarrow B \rightarrow C_{ReLU}(4,2,1024) \rightarrow B \rightarrow FC(64)$.

\noindent{\bf MDN:} 
The \textit{MDN}'s  input is a $28 \times 28 \times 512$ gray-level feature embedding from [48]. It predicts the parameters of $8$ Gaussian components, namely
$8$ means of size $64$, and $8$ mixture weights. We use a fixed spherical variance $\sigma$ equal to $0.1$. The \textit{MDN} network is constructed as follows: $28 \times 28 \times 512 \rightarrow C_{ReLU}(5,1,384) \rightarrow B \rightarrow C_{ReLU}(5,1,320) \rightarrow B \rightarrow C_{ReLU}(5,1,288) \rightarrow B \rightarrow C_{ReLU}(5,2,256) \rightarrow B \rightarrow C_{ReLU}(5,1,128)\rightarrow B\ \rightarrow FC(4096) \rightarrow FC(8 \times 64+8)$.

\noindent{\bf Decoder:} 
During training, the \textit{decoder}'s input is the embedding of size $64$ generated by the \textit{encoder}. At test time, the input is a $64$-dimensional latent code sampled from one of the  Gaussian components of the \textit{MDN}. The \textit{decoder} generates a vector $\vect B$ of unary terms of size $(32 \times 32 \times 2)$. It consists of operations of bilinear up-sampling and convolutions. During  training, we learn four feature maps $f_{1}$, $f_{2}$, $f_{3}$ and $f_{4}$ of the gray-level image of sizes 
$4 \times 4 \times 1024$, $8 \times 8 \times512$,  $16 \times 16 \times 256$ and $32 \times 32 \times 128$, respectively. We concatenate these representations with the ones learned by the decoder, after every up-sampling operation, \ie, $f_{1}$ is concatenated with the decoder's representation after the first up-sampling, $f_{2}$ after the second upsampling and so on. The architecture is described as follows: $1 \times 1 \times 64 \rightarrow U(4)  \rightarrow C_{ReLU}(4,1,1024) \rightarrow B\rightarrow U(2) \rightarrow C_{ReLU}(5,1,512) \rightarrow B \rightarrow U(2) \rightarrow C_{ReLU}(5,1,256) \rightarrow B \rightarrow C_{ReLU}(5,1,128) \rightarrow B \rightarrow U(2) \rightarrow C_{ReLU}(5,1,64) \rightarrow B \rightarrow C_{ReLU}(5,1,2) \rightarrow (32 \times 32 \times 2)$.

\noindent{\bf Structure Encoder:} 
The structure encoder is used to learn a $512$-dimensional embedding for every node in the down-sampled gray-level image of size $32 \times 32 \times 1$. It consists of stacked convolutional layers: $(32 \times 32 \times 1)  \rightarrow  C_{ReLU}(5,1,16) \rightarrow B  \rightarrow  C_{ReLU}(5,1,32)  \rightarrow  B\rightarrow C_{ReLU}(5,1,64) \rightarrow B \rightarrow  C_{ReLU}(5,1,128) \rightarrow B \rightarrow C_{ReLU}(5,1,256) \rightarrow B  \rightarrow  C_{ReLU}(5,1,512) \rightarrow 512 \times (32 \times  32)$. 

\noindent{\bf G-CRF Layer:} 
The G-CRF is implemented as an additional layer that has as input the unary term $\vect B$ and the embedding matrix $\vect{\mathcal{A}}$ produced by the structure encoder. During the forward pass, it outputs a color field of size $32 \times 32 \times 2$ by solving the linear system in Eq. 5 of the main paper for the channels $a$ and $b$ separately. During the backward pass, it generates the gradient of the loss with respect the unary term and the the embedding matrix respectively. To compute the gradient of $\vect B$, we  solve the linear system in Eq.~\ref{eq:grad_B}. The gradient of $\vect{A}$ is given in Eq.~\ref{eq:grad_A}.

\section{Additional Results}
\label{sec:add_results}

\begin{figure*}[!h]
    \centering
     \vspace{-1cm}
  \includegraphics[width=5in, height=3.2in]{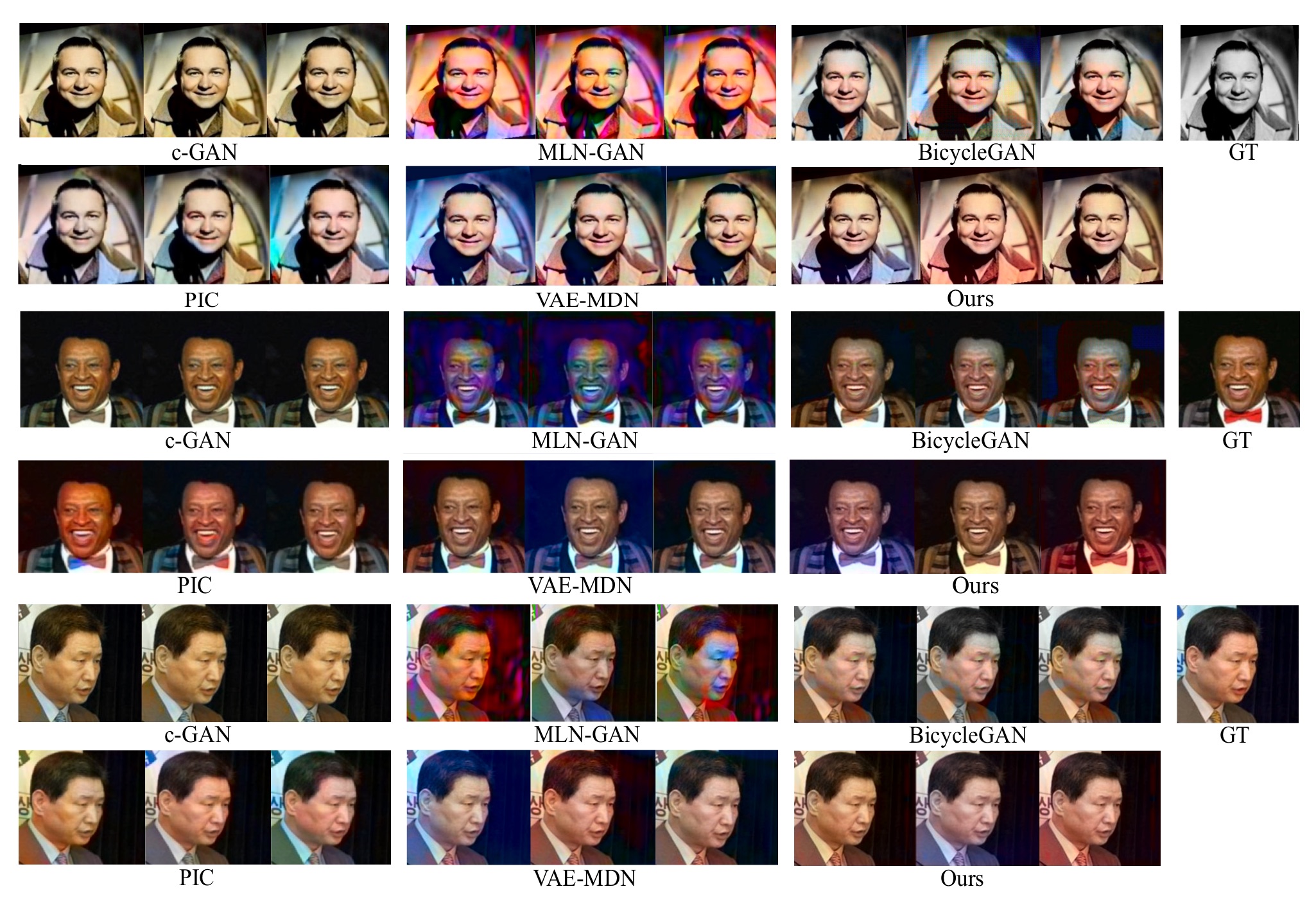}
  \vspace{-0.7cm}
  \caption{Qualitative comparison of our results with the baselines on LFW.}
   
\label{fig:lfw_results}
\end{figure*}


\begin{figure*}[!htbp]
    \centering
    \vspace{-0.7cm}
  \includegraphics[width=5in, height=3.4in]{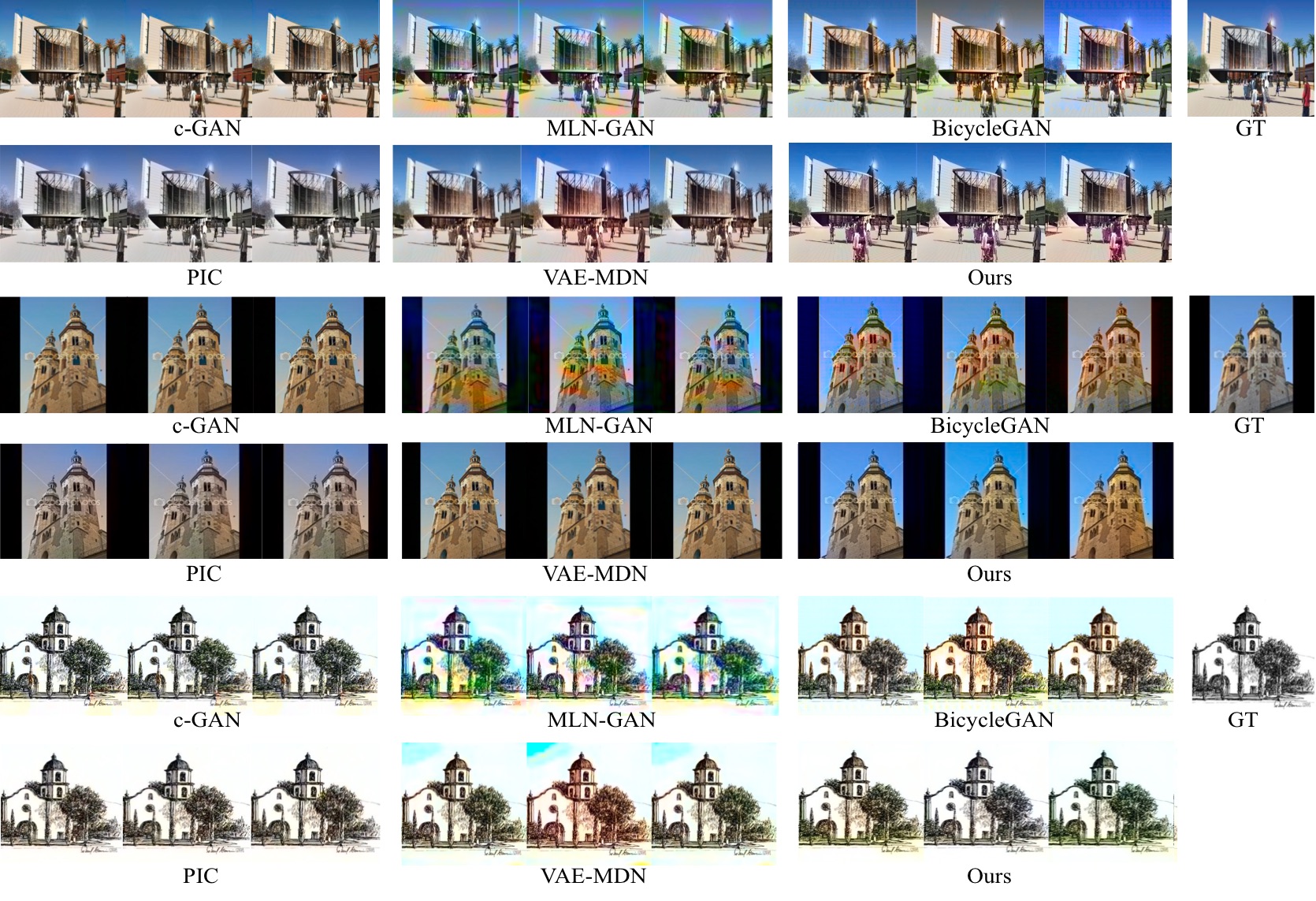}
\vspace{-0.7cm}
\caption{Qualitative comparison of our results with the baselines on LSUN.}
   
\label{fig:sun_results}
\end{figure*}


\begin{figure*}[!htbp]
    \centering
  \includegraphics[width=5in, height=3.2in]{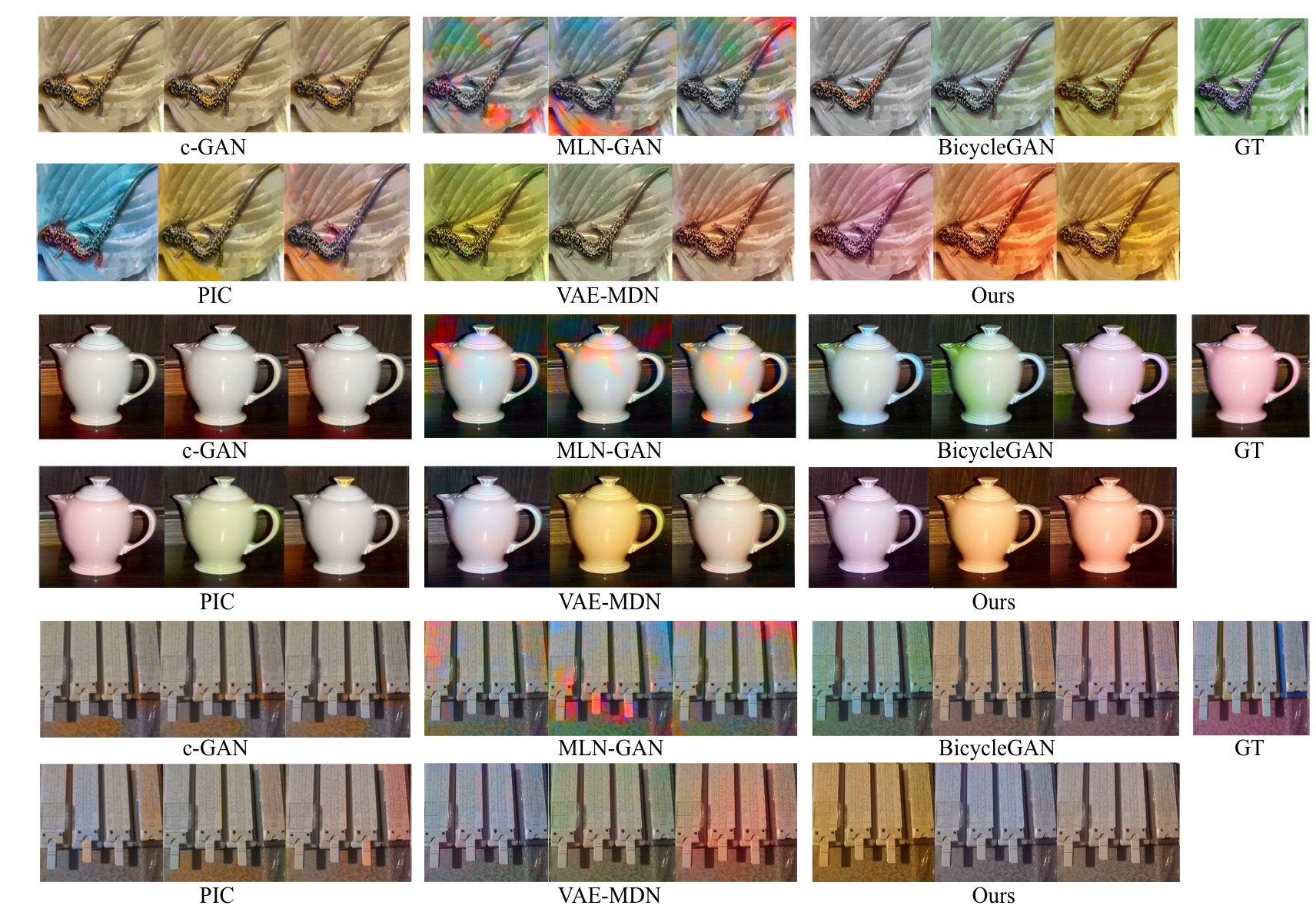}
  \vspace{-0.7cm}
    \caption{Qualitative comparison of our results with the baselines on the ILSVRC-2015 dataset.}   
\label{fig:imagenet_results}
\end{figure*}

\begin{figure}[!htbp]
    \centering \hfill   
    \vspace{-0.7cm}
    \includegraphics[width=5in, height=3.in]{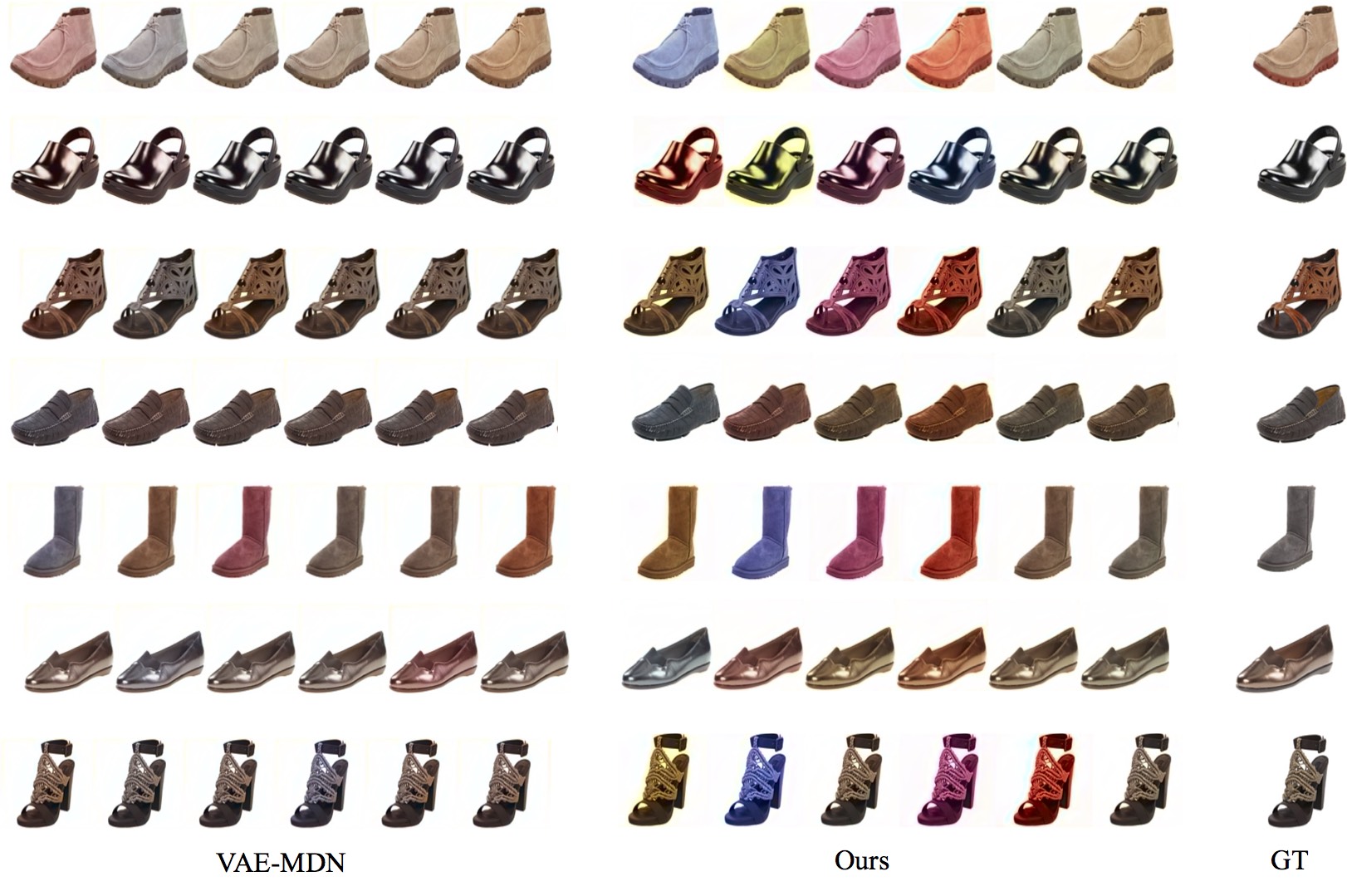}   
    \vspace{-0.7cm}          
    \caption{Qualitative comparison of our results with the baselines on the Shoes dataset~\cite{shoes_dataset}. }
    \label{fig:classical_vae}
    
    \label{fig:shoes}
\end{figure}

\newpage
\section{Results on ImageNet Dataset}
\label{sec:ImageNetresults}
\vspace*{-20pt}
\begin{figure*}[!htbp]
    \centering
    \vspace{-0.2cm}
     \begin{subfigure}[b]{\textwidth}
        \includegraphics[width=5in, height=1.3in]{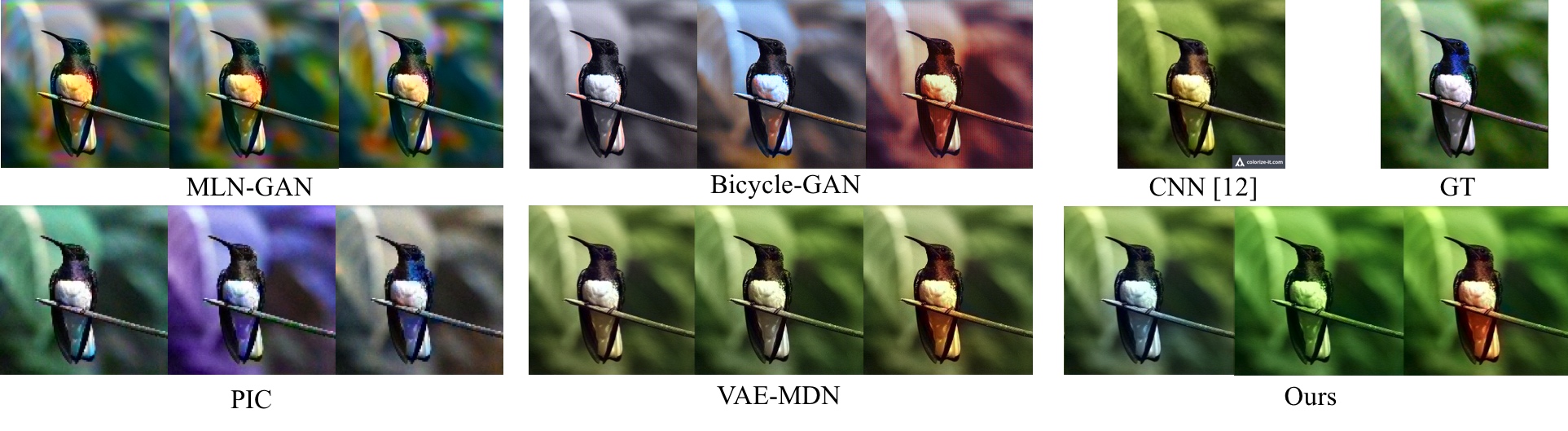}
        \label{fig:lezard}
    \end{subfigure}
\vspace{-0.2cm}
    \begin{subfigure}[b]{\textwidth}
        \includegraphics[width=5in, height=1.3in]{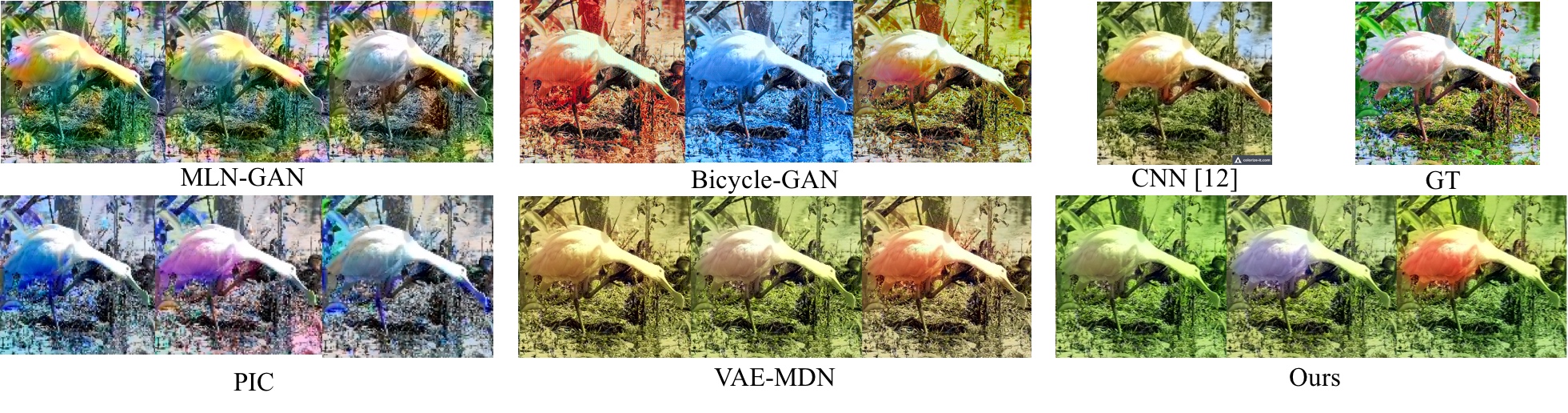}
        \label{fig:rex_ritter}
    \end{subfigure}
    \vspace{-0.2cm}
     \begin{subfigure}[b]{\textwidth}
        \includegraphics[width=5in, height=1.3in]{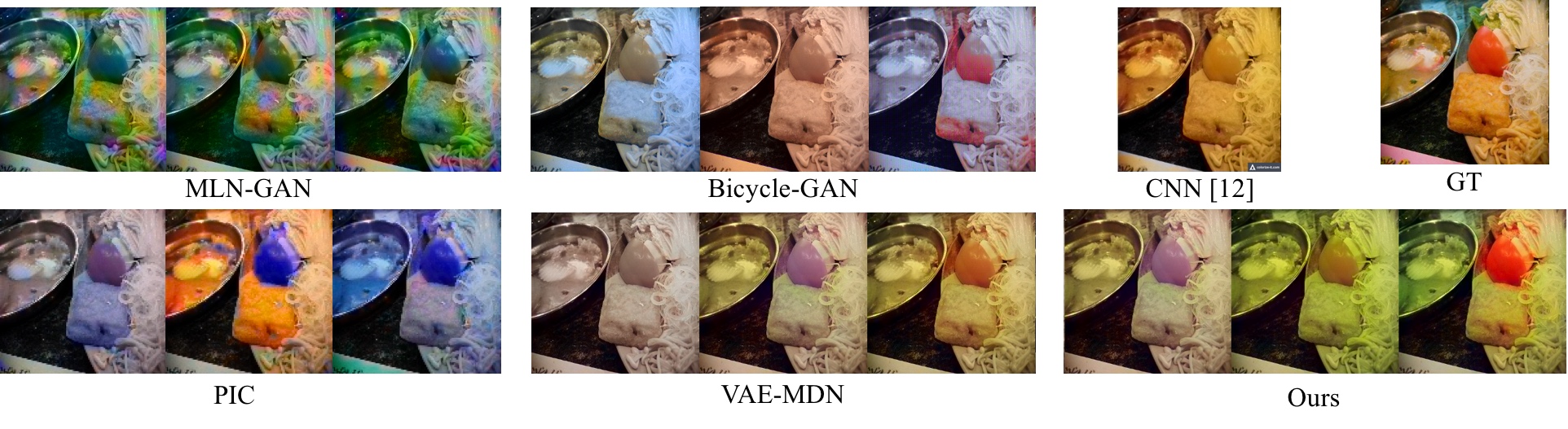}
        \label{fig:rex_ritter}
    \end{subfigure}
\vspace{-0.2cm}
       \begin{subfigure}[b]{\textwidth}
        \includegraphics[width=5in, height=1.3in]{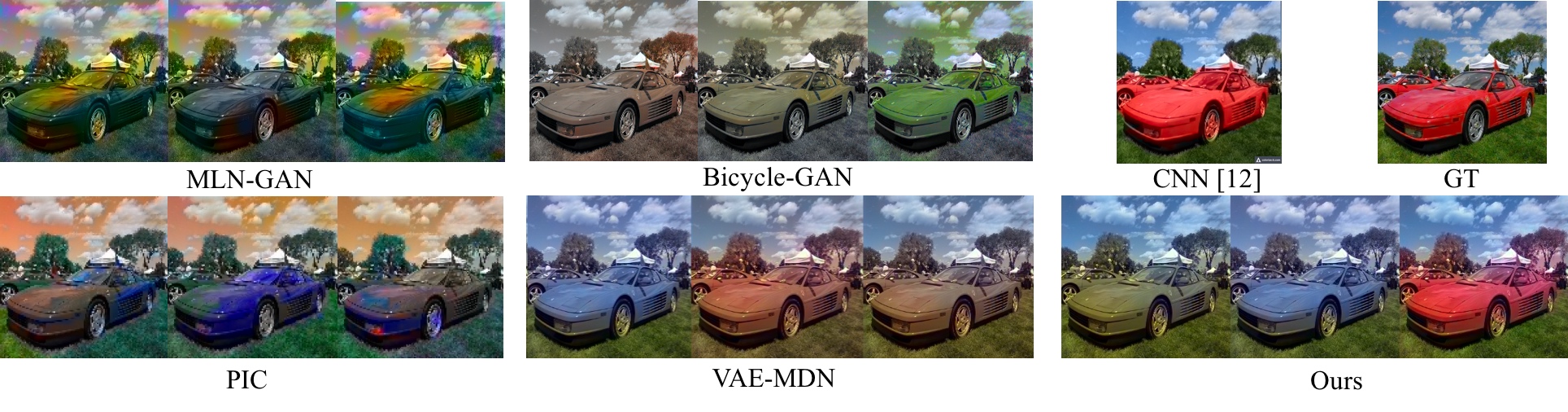}
        \label{fig:police}
    \end{subfigure}
    
\vspace{-0.2cm}
 \begin{subfigure}[b]{\textwidth}
	\includegraphics[width=5in, height=1.3in]{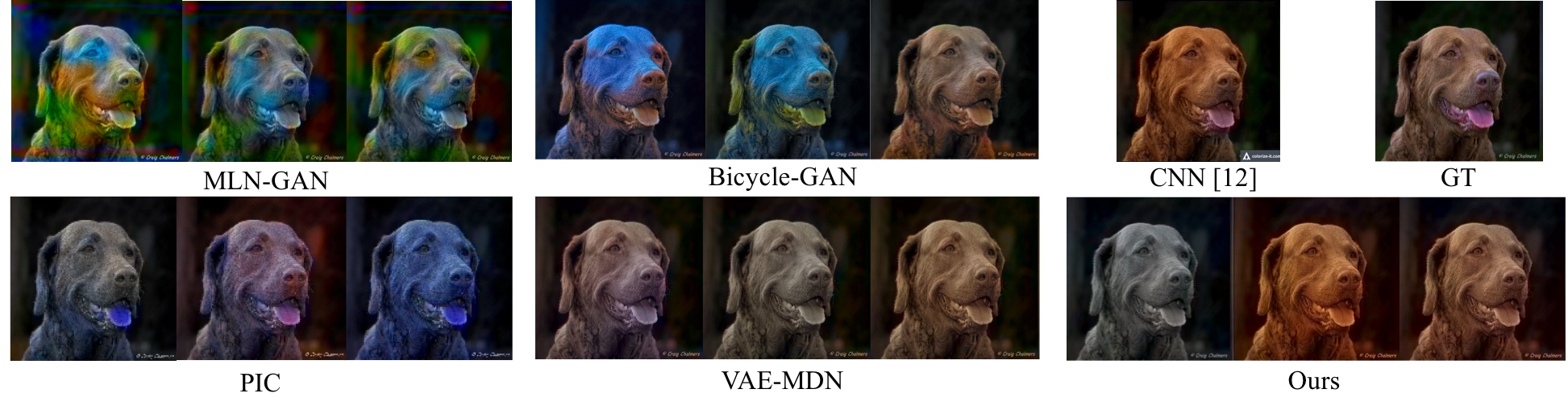}
	\label{fig:arbitre}
\end{subfigure}    
    \vspace{-1cm}
    \caption{Qualitative comparison of our results with the baselines on the ImageNet dataset.}
\label{fig:imagenet_new_results}

\end{figure*}

\newpage

\section{Beyond Colorization: G-CRF for Structured Generative Models}
\label{sec:Beyond Colorization}
\begin{figure*}[!htbp]
    \centering
    \begin{subfigure}[b]{0.32\textwidth}
        \includegraphics[width=\textwidth]{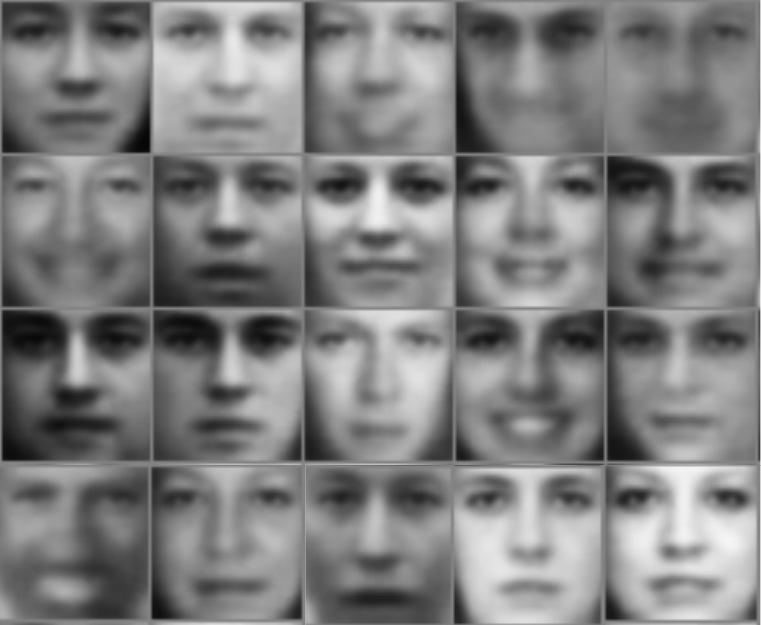}          
        \caption{\footnotesize{Classical VAE }}
        \label{fig:classical_vae}
    \end{subfigure}
    \begin{subfigure}[b]{0.32\textwidth}
        \includegraphics[width=\textwidth]{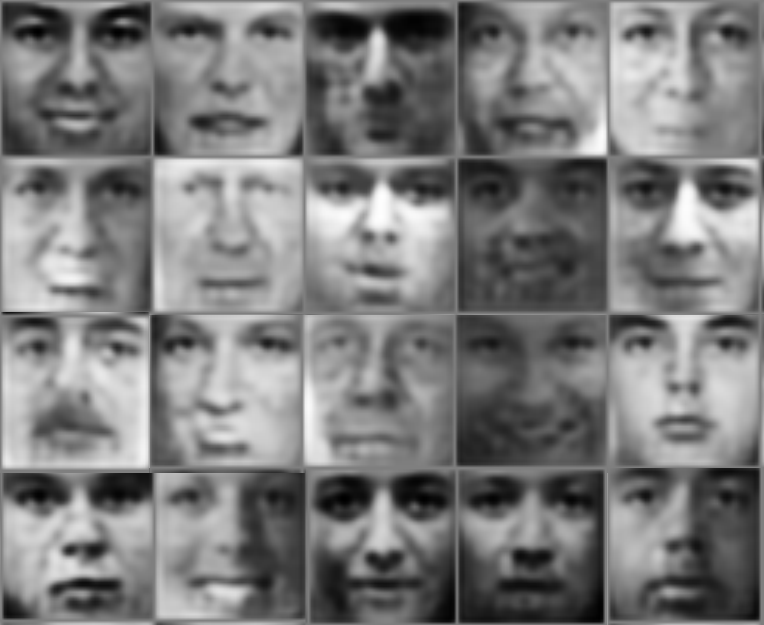}             
\caption{\footnotesize{Feature Consistent VAE} }       
        \label{fig:feature_vae}
    \end{subfigure} 
    \begin{subfigure}[b]{0.32\textwidth}
        \includegraphics[width=\textwidth]{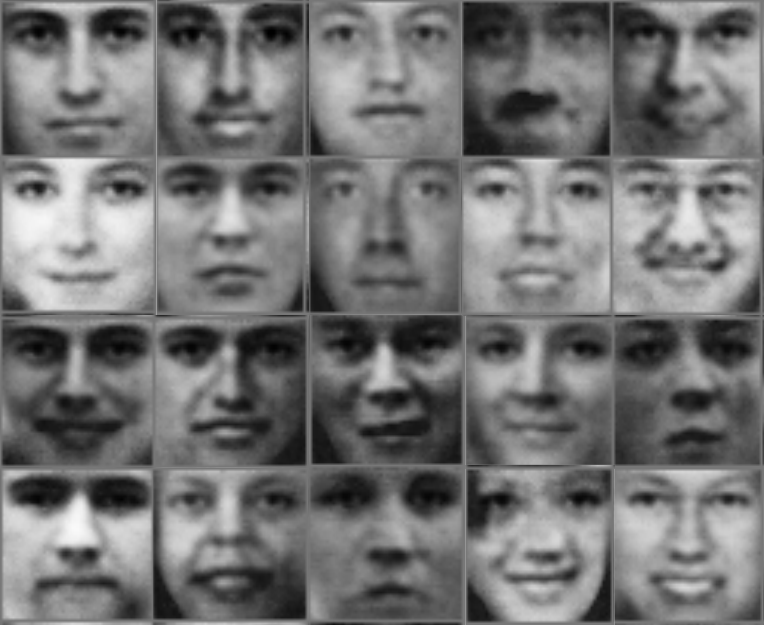}      
        \caption{\footnotesize{Structured VAE (Ours)}}
        \label{fig:s_vae}
    \end{subfigure}\hfill 
     \vspace{-0.2cm}
     \caption{ {\small Randomly generated samples from a classical VAE~\cite{kingma2013}, a deep feature consistent VAE~\cite{hou2017deep} and our structured output space feature consistent VAE, trained on TFD. } }
    \label{fig:vae}
\end{figure*}

\begin{figure*}[!htbp]
    \centering \hfill   
    \begin{subfigure}[b]{0.32\textwidth}
        \includegraphics[width=\textwidth]{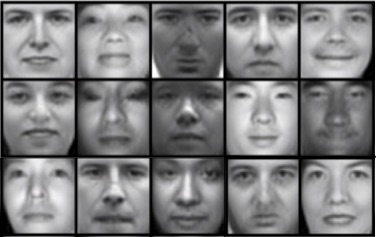}             
        \caption{Classical BEGAN }
        \label{fig:classical_vae}
    \end{subfigure}\hfill   
    \begin{subfigure}[b]{0.32\textwidth}
        \includegraphics[width=\textwidth]{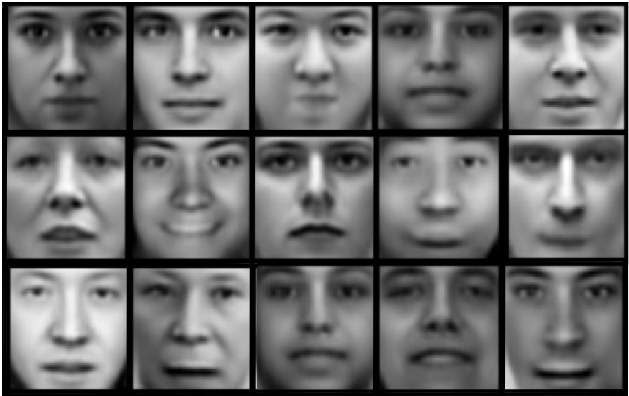}       
\caption{Structured BEGAN }        
        \label{fig:feature_vae}
    \end{subfigure} \hfill   
    \vspace{-0.2cm}
     \caption{{\small Randomly generated samples from a classical BEGAN~\cite{berthelot2017began} and our BEGAN with a structured output space discriminator trained on TFD. } }
    \label{fig:bgan}
\end{figure*}

Beyond colorization, we explore the effect of endowing two different generative models, namely variational auto-encoders and Boundary Equilibrium Generative Adversarial Network (BEGAN)~\cite{berthelot2017began} with a structured output space through our G-CRF formulation. We show the results in  Fig.~\ref{fig:vae} and Fig.~\ref{fig:bgan} using the Toronto Face Dataset (TFD)~\cite{susskind2010toronto}. Quantitative results are reported in Tab.~\ref{tab:generative_model} using (1) KL divergence between the distributions of  generated and real data, (2) sharpness by gradient magnitude, (3) by edge width and (4) by variance of the Laplacian. The results are normalized with respect to real data measurements.

For the variational auto-encoder model, The G-CRF is added on top of the decoder. Additionally, the reconstruction loss is augmented with the feature loss from~\cite{hou2017deep}. We compare our results with the ones obtained from a classical VAE and a VAE trained with the feature loss without the G-CRF layer. Fig~\ref{fig:vae} shows that our model results in sharper, higher quality and more diverse faces. 

For the BEGAN model, we add our G-CRF layer on top of the discriminator. Hence, the model implicitly penalizes generated samples which have different statistics than the real samples, at the output layer level.  In Fig.~\ref{fig:bgan}, we compare our results with the classical BEGAN for the hyper-parameter gamma set to .5 after approx.~120,000 iterations. We observe our model to generate diverse and better quality samples.
\vspace{-5pt}

\begin{table}[H]
\centering
\caption{Results of the CRF extension for generative models}
\label{tab:generative_model}
\setlength\tabcolsep{4pt}
\begin{tabular}{cccccc} \toprule
\emph{}  & \emph{\footnotesize KLD} & \emph{\footnotesize Gradient}   & \emph{\footnotesize Edge width} & \emph{\footnotesize Laplacian}   \\ \midrule
\footnotesize Classical VAE & \footnotesize 0.33 & \footnotesize 66.5\% &  \footnotesize 52.65\% & \footnotesize 16.3\% \\ 
\footnotesize Feature Consistent VAE & \footnotesize 0.16 & \footnotesize 91.5\% & \footnotesize 72.9\% & \footnotesize 30.17\% \\ 
\footnotesize S-VAE (Ours)& \footnotesize $\bm{0.13}$ & \footnotesize $\bm{97\%}$. & \footnotesize $\bm{95.5\%}$ & \footnotesize  $\bm{92.\%}$   \\  \midrule
 \footnotesize  BEGAN &\footnotesize  0.09 & \footnotesize 98.54 \% & \footnotesize 99.4 \%  &  \footnotesize 96.37\%   \\ 
  \footnotesize  S-BEGAN (Ours) & \footnotesize  $\bm{0.07}$ & \footnotesize $\bm{99.1\%}$ & \footnotesize  $\bm{100. \%}$ &\footnotesize  $\bm{98.76\%}$   \\ 
 \bottomrule
\end{tabular}
\end{table}

\end{document}